\documentclass[10pt,twocolumn,letterpaper]{article}

\usepackage{btas}
\usepackage{times}
\usepackage{epsfig}
\usepackage{graphicx}
\usepackage{amsmath}
\usepackage{amssymb}

\usepackage{btas}
\usepackage{times}
\usepackage{epsfig}
\usepackage{graphicx}
\usepackage{amsmath}
\usepackage{amssymb}
\usepackage{gensymb}
\usepackage{subcaption}
\usepackage{url}

\usepackage{array}
\usepackage{multirow}
\usepackage{textcomp}
\usepackage{array}
\usepackage{colortbl}
\graphicspath{{./graphics/}}
\usepackage[table]{xcolor}
\usepackage{booktabs}

\usepackage[norule,symbol,perpage]{footmisc}

\definecolor{myblue}{rgb}{0,0,1}
\definecolor{myred}{rgb}{0.8, 0, 0}
\definecolor{mygreen}{rgb}{0, 0.6, 0}

\usepackage{subcaption}

\btasfinalcopy 

\ifbtasfinal\fi
\begin{document}

\title{Perception of Image Features in Post-Mortem Iris Recognition: \\Humans vs Machines}

\author{Mateusz Trokielewicz\\
Biometrics and Machine Intelligence Lab\\
Research and Academic Computer Network\\
Kolska 12, 01045 Warsaw, Poland\\
{\tt\small mateusz.trokielewicz@nask.pl}
\and
Adam Czajka\\
Department of Computer Science and Engineering\\
University of Notre Dame\\
Notre Dame, IN 46556, USA\\
{\tt\small aczajka@nd.edu}
\and
Piotr Maciejewicz\\
Department of Ophthalmology\\
Medical University of Warsaw\\
Lindleya 4, 02005 Warsaw, Poland\\
{\tt\small piotr.maciejewicz@wum.edu.pl}
}

\maketitle
\thispagestyle{empty}

\begin{abstract}
Post-mortem iris recognition can offer an additional forensic method of personal identification. However, in contrary to already well-established human examination of fingerprints, making iris recognition human-interpretable is harder, and therefore it has never been applied in forensic proceedings. There is no strong consensus among biometric experts which iris features, especially those in iris images acquired post-mortem, are the most important for human experts solving an iris recognition task. This paper explores two ways of broadening this knowledge: (a) with an eye tracker, the salient features used by humans comparing iris images on a screen are extracted, and (b) class-activation maps produced by the convolutional neural network solving the iris recognition task are analyzed. Both humans and deep learning-based solutions were examined with the same set of iris image pairs. This made it possible to compare the attention maps and conclude that (a) deep learning-based method can offer human-interpretable decisions backed by visual explanations pointing a human examiner to salient regions, and (b) in many cases humans and a machine used different features, what means that a deep learning-based method can offer a complementary support to human experts. This paper offers the first known to us human-interpretable comparison of machine-based and human-based post-mortem iris recognition, and the trained models annotating salient iris image regions.  
\end{abstract}

\section{Introduction}
\label{sec:introduction}
Recent research has unveiled the potential that the iris might be useful in post-mortem identification and verification of humans \cite{TrokielewiczPostMortemICB2016, TrokielewiczPostMortemBTAS2016, BolmeBTAS2016, TrokielewiczTIFS2018}. These studies, conducted in both the mortuary, cold-storage conditions, as well as in uncontrolled outside environment, have shown that correct matches can be obtained with cadaver irises even three weeks after death. However, existing iris matchers are weakly suited for this task, with error rates growing with increased time horizon since subject's death. There are also no human-interpretable post-mortem iris recognition methods reported in the literature to help human examiners in their work. If post-mortem iris biometrics can be successfully implemented, it could be a valuable addition to the forensic expert's set of methods for identification, proving useful in cases when other methods, such as DNA or dental records, are unavailable or difficult to apply. It is easy to imagine a scenario of a hypothetical natural disaster victim search, when a fast positive identification can free up valuable resources of emergency response teams and let them proceed with shorter delay.
     
At the same time, simply providing a machine-backed decision on to whom the iris might belong would not be considered sufficient during courthouse proceedings, similarly to the case of fingerprints, where the automated fingerprint identification systems (AFIS) serve only as assistance to the human expert, who is making the final decision. Such use case drives the motivation of this work, in which we propose an algorithm incorporating deep convolutional neural network (DCNN) for cadaver iris recognition that, in addition to its class-wise prediction, also offers a visualization of the salient regions used by a classifier. Furthermore, we compare attention maps generated by the neural network with attention maps obtained from human subjects with the use of an eye tracker device, to gain insight into how differently a machine and humans perform in this task, which iris regions they deem important, and whether the two methods can complement each other.

With this paper, we try to deliver answers for the following two questions:

\begin{enumerate}
	\item[Q1.] Which iris regions contribute the most to the class-wise prediction made by a DCNN trained for iris recognition?
	\item[Q2.] How do the DCNN-generated attention maps compare to the maps obtained from an eye tracker device recording human's eye gaze during iris recognition task?
\end{enumerate}

To our knowledge, this is the first work analyzing differences between attention to iris features in a DCNN classifier and in human subjects. We also make the trained DCNN classifier, annotating salient iris features, available along with this paper at: \url{http://zbum.ia.pw.edu.pl/EN/node/46}.

\section{Related work}
\label{sec:related}

\subsection{Post-mortem iris recognition}

Sansola \cite{BostonPostMortem} used IriShield M2120U iris recognition camera together with IriCore matching software in experiments involving 43 subjects who had their irises photographed at different post-mortem time intervals. Depending on the post-mortem interval, the method yielded 19-30\% of false non-matches and no false matches. Saripalle \etal \cite{PostMortemPigs} used {\it ex-vivo} eyes of domestic pigs with a conclusion that irises are slowly degrading after being taken out of the body, and lose their biometric capabilities 6 to 8 hours after death. Ross \cite{RossPostMortem} drew some conclusions on the development of corneal opacity and fadeout of the pupillary and limbic boundaries in post-mortem samples. Trokielewicz \etal have shown that the iris can still serve as a biometric identifier for 27 hours after death \cite{TrokielewiczPostMortemICB2016}, even with the existing iris matchers. Later, they showed that correct matches can still be expected after 17 days since a subject's death \cite{TrokielewiczPostMortemBTAS2016}. A database of 1330 near infrared and visible light post-mortem iris images acquired from 17 cadavers was offered to the scientific community. Recent study by Trokielewicz \etal \cite{TrokielewiczTIFS2018}, employing images collected up to 34 days post-mortem from 37 cadavers, shows that iris recognition occasionally works even 21 days since a subject's demise. 

Bolme \etal \cite{BolmeBTAS2016} pioneered the analysis of how fast faces, fingerprints and irises are losing their biometric capabilities during human decomposition in natural, outdoor environment, and in different weather conditions. The authors showed that the irises degraded quickly regardless of the temperature, typically becoming useless only a few days after placement. A recent paper by Sauerwein \etal \cite{Sauerwein_JFO_2017} followed these experiments, showing that irises stay readable for up to 34 days after death, when cadavers were kept in outdoor conditions during winter. Their readability, however, was assessed by human experts, and not by specialized iris recognition algorithms.

Some advancements have recently been made in automated post-mortem iris biometrics, with an algorithm for cadaver iris image segmentation that is said to effectively learn specific, post-mortem deformations of the iris texture, and successfully exclude them during segmentation proposed by Trokielewicz \etal \cite{TrokielewiczIWBF2018}, as well as a method for detecting iris images coming from post-mortem subjects, which correctly detects almost 99\% of the cadaver sample presentations \cite{TrokielewiczColdPAD_BTAS2018}.

\subsection{Applications of convolutional neural networks for iris recognition}

Over the last few years, several deep learning-based approaches to iris recognition have been proposed as an alternative to typical methods employing conventional, hand-crafted iris feature representations, such as those based on the works of John Daugman \cite{Daugman1993, Daugman2007NewMethods}. Minaee \etal \cite{MinaeeCNNforIrisRec2016} extracted features from the entire eye region using a pre-trained network based on the {\it VGG-Net} architecture \cite{VGGSimonyanCNNsForRecognition2014} with no fine-tuning, and an SVM applied as a classifier. Their solution, tested on CASIA-Iris-1000 database (20,000 iris images from 1,000 subjects) and IIT Delhi database (2,240 images from 224 subjects), reached 88\% and 98\% recognition rates, respectively. 

Gangwar and Joshi \cite{DeepIrisNet2016Gangwar} introduced {\it DeepIrisNet}, constituting two convolutional architectures built specifically for the purpose of iris recognition, one being a typical, pyramid-like structure of stacked convolutional layers, and the second being an inception-style network coupled with stacked convolutional layers. These were trained in a closed-set scenario, then the softmax layer was removed and the output from the last dense layer was extracted to provide a 4096-dimensional vector of iris features, compared using Euclidean distance. An equal error rate (EER) of 1.82\% was reported. 

Liu \etal \cite{DeepIrisLIU2016} introduced a {\it DeepIris} network designed for iris images coming from two different sensors, with different resolution, quality, etc. The solution depends on pairs of features that are learnt from the data. The experiments involved a CNN architecture comprising several convolutional layers, trained and tested on subject-disjoint subsets of Q-FIRE and CASIA cross-sensor datasets. EER = 0.15\% is reported. 

Zhao and Kumar \cite{ZhaoDeepIrisICCV2017} proposed a fully-convolutional network architecture for iris masking and representation, trained with the use of a triplet loss function with bit-shifting and iris masking. The approach employs binarization of the network output and additional masking of the `less reliable' bits in the feature map, similarly to the concept of ignoring fragile bits in iris code \cite{Hollingsworth_BTAS_2009}. This method gave EERs of 0.73\% and 0.99\% for the IITD and ND-Iris-0405 datasets, and 2.28\% and 3.85\% for more challenging WVU Non-ideal and CASIA.v4-distance datasets.

Nguyen \etal \cite{IrisCNNsOffTheShelfNguyen2018} explored off-the-shelf features obtained from selected modern CNN architectures, coupled with a multi-class SVM classifier. The best performing models include DenseNet (best), ResNet, and Inception. Good recognition rates are reported, nearing 99\% for the two databases used in the paper, namely the LG2200 and CASIA-Iris-Thousand, albeit for a closed-set experiment.

\begin{figure*}[t]
	\centering
	\includegraphics[width=0.43\textwidth]{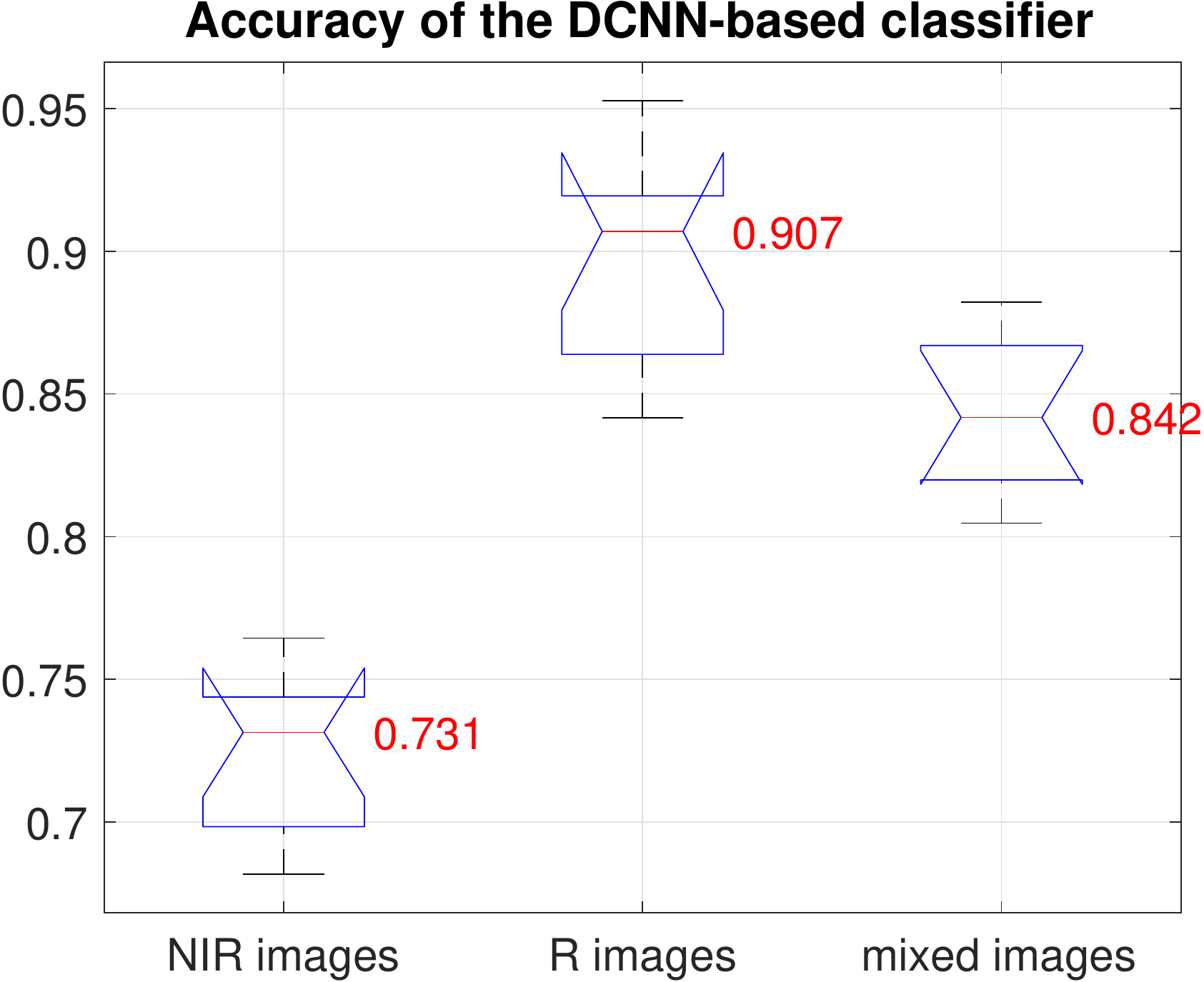}\hskip8mm
		\includegraphics[width=0.42\textwidth]{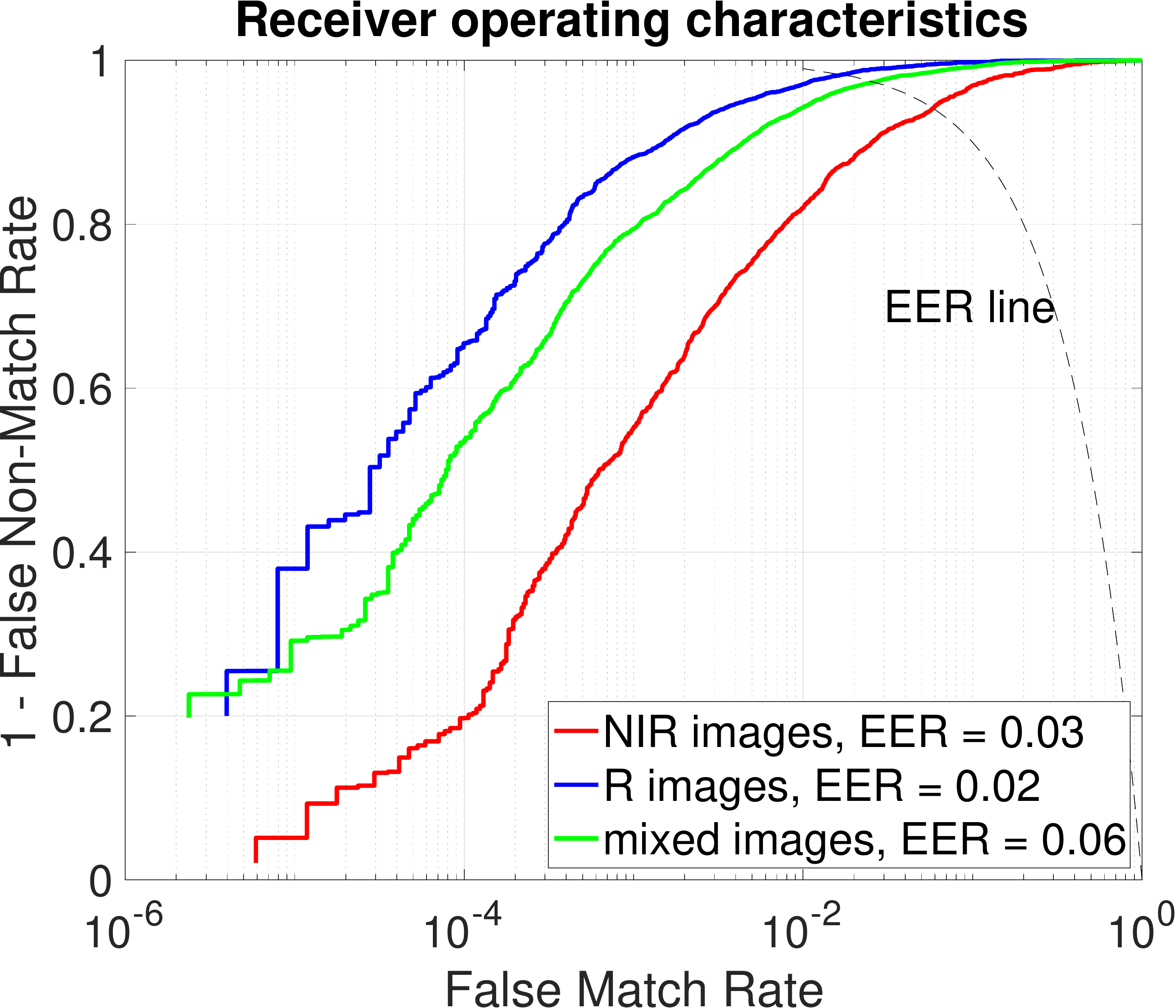}
	\caption{Performance of our DCNN-based classifier in terms of: classification accuracy \textbf{(left)} and Receiver Operating Characteristic (ROC) curves with Equal Error Rates \textbf{(right)}, when trained on near infrared (NIR), red channel (R) images, and NIR+R (mixed) images.}
	\label{fig:boxplots_cold}
\end{figure*} 

\section{Datasets of cadaver iris images}
\label{sec:data}
In this work we take advantage of the two publicly available, subject-disjoint datasets of iris images collected from cadaver eyes: Warsaw-BioBase-PostMortem-Iris v1 and v2\footnote{\url{http://zbum.ia.pw.edu.pl/EN/node/46}}, which combined contain 1200 near infrared (NIR) images and 1787 visible light images obtained from 37 subjects. These samples were collected in mortuary conditions over a period of time reaching up to 34 days post-mortem, in multiple sessions across different time horizons after death. Each cadaver eye was imaged multiple times in several (from 2 to 13) acquisition sessions. A total of 72 eyes are represented in the data, since two subjects had only one of their eyes photographed. In addition, data for one of the classes had to be removed from analysis as it was only represented by a single NIR sample. Thus, the final database used in this study consists of 1199 NIR samples and 1780 visible-light samples, representing 71 distinct eyes. Since left and right irises are different, we assume that each eye represents a separate identity, or class. For the purpose of both training the machine classifier as well as experiments involving human subjects, the images were manually segmented with circular approximations of the iris boundaries and cropped to square.

\section{DCNN-based iris classifier}
\label{sec:trainingAndEval}
For the purpose of constructing our classifier, we take advantage of the {\it VGG-16} model \cite{VGGSimonyanCNNsForRecognition2014}, which is pre-trained on the ImageNet database of natural images. The number of network outputs was adapted to the number of individual eyes, and fine-tuning was performed with the Warsaw-BioBase-PostMortem-Iris v1 and v2 datasets comprising images of 71 eyes. Such approach has been found by many researchers as the best way to adapt a CNN to a new domain, with high chances to get a model that presents sufficient generalization capabilities. 10 independent train/test data splits were created by randomly assigning 80\% of the data in each class to the training subset, and the remaining 20\% of the data to the testing subset. The training took 30 epochs in each of the train/test split, and involved stochastic gradient descent with momentum $m=0.9$, learning rate of $0.0001$, and mini-batch size of $16$. Experiments were repeated three times with different types of iris image data: near-infrared images (NIR), red-channel images extracted from high-resolution RGB images (R), and with a combined dataset of both types of data (mixed).

During testing, softmax outputs were utilized to plot the receiver operating characteristic (ROC) curves for our DCNN-based classifier for three types of  training data. This, together with the classification accuracies (proportions of samples being correctly classified by the network to the overall number of test samples in a given train/test split) and equal error rates are shown in Fig. \ref{fig:boxplots_cold}. The model trained with R images is performing best, with EER as low as 1.74\% and an average classification accuracy of 90.7\%, which can be attributed to better quality offered by these images, at least for early acquisition sessions. Also, most of the subjects in the experimental database (29 out of 37) had lightly-colored eyes (\ie gray, blue, or light-green), which are known to offer better visibility of the iris texture under visible light illumination. This prevalence of light-colored eyes can lead to an overestimation of the classification accuracy when compared to a more diverse population. The model trained with NIR data performs worse (EER=5.73\% and accuracy of 73.1\%), but the model employing both kinds of data is only slightly worse than the R model, offering EER=2.5\% and an accuracy of 84.2\%. These results allow to conclude that the DCNN model offers a decent post-mortem iris recognition tool that will be used in the core component of this research presented in the next Section. Note that the observed results, on average worse than usually observed in iris recognition, correspond to a challenging biometric task: {\it post-mortem} iris recognition.

\section{Humans vs machines}
\label{sec:CAMAndEyetracking}
In this Section, we employ two methods, namely the Grad-CAM algorithm described in Sec. \ref{sec:related}, and the eye tracking technique, to obtain attention maps highlighting regions of the iris image considered important when making the decision by the machine and by humans.

\subsection{Machine-based attention maps}
\label{sec:CAM}
Basic DCNN's designs do not provide a human-interpretable explanation for their decisions, which makes them unsuitable for assisting human experts in a courtroom scenario, because a softmax output cannot be expected to convince the jury of a person's innocence or guilt. 

For identification of discriminative image regions, decisive for the model prediction, class activation mapping (CAM) techniques have been proposed, first introduced by Zhou \etal \cite{CAMZhou}. The authors achieve this by removing fully-connected layers and replacing them with global average pooling layers followed only by a softmax layer. As a result, image regions that are important for discrimination are highlighted with a heatmap. Selvaraju \etal improved Zhou's method with Grad-CAM \cite{GradCAMSelvaraju}, which does not require any changes to the network's architecture and yields coarse heat-maps highlighting the regions that contribute the most to the model's prediction.

By using these methods with our DCNN cadaver iris classifier, we are able to provide the human expert more knowledge on \textit{why} a probe iris is assigned to a given class in addition to stating \textit{which} class it most likely belongs to. 


To obtain machine-based attention maps, we take advantage of the method introduced in \cite{GradCAMSelvaraju}, by training the classification network in the same manner as described in Section \ref{sec:trainingAndEval}. An adapted code from \cite{GradCAMCode} is used for the implementation in Keras/Tensorflow environment \cite{Keras, tensorflow2015}. A modified training procedure is employed here, where the subset of data that was used in the gaze-tracking part of the experiments constitutes the testing subset, and the remaining data is assigned to the training subset. The training samples are segmented manually.

\begin{figure}[t]
	\centering
	\includegraphics[width=0.49\textwidth]{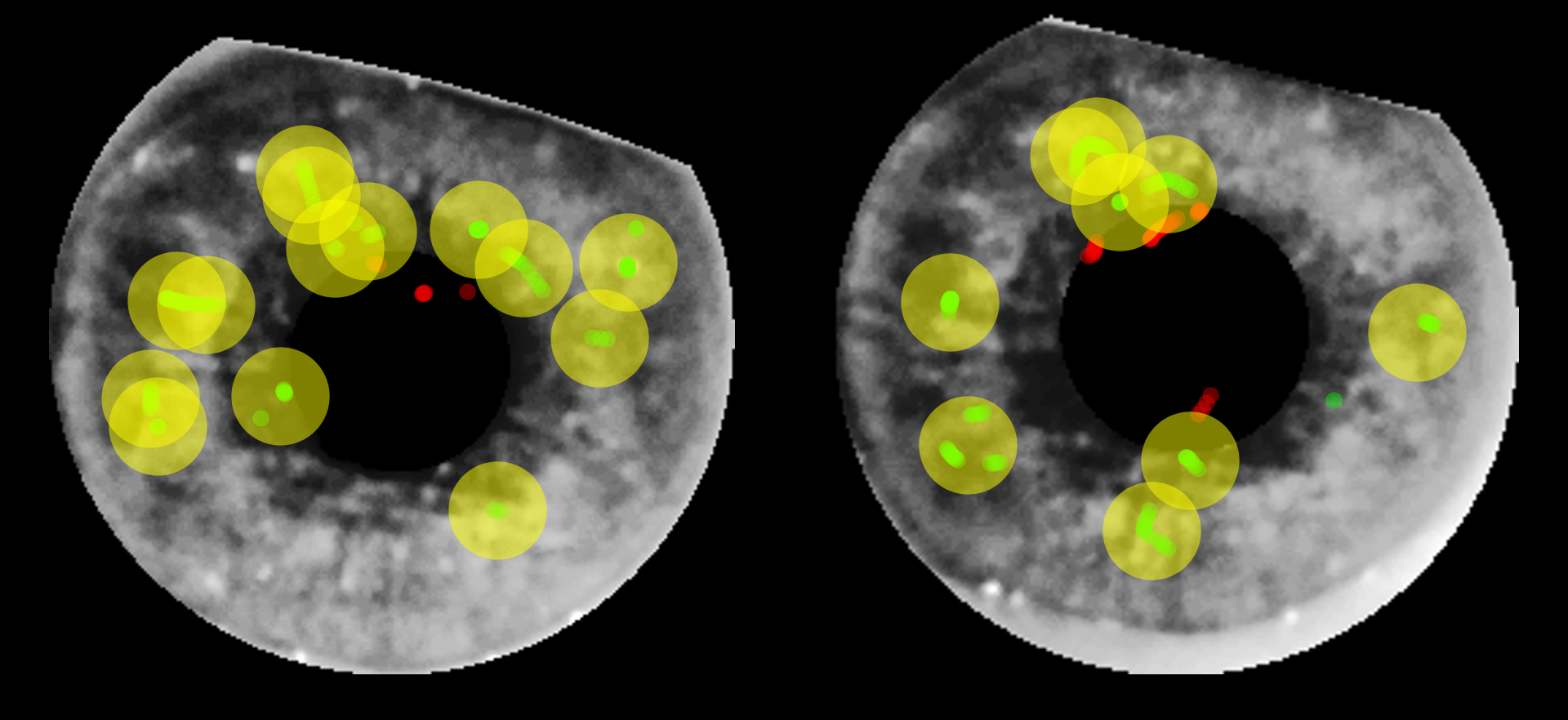}
	\caption{Example attention maps for the same iris image pair coming from cadaver eyes, recorded during the experiment. Green and red dots represent the raw gaze fixation points (within, and outside of the iris, respectively), whereas the yellow circles denote the averaged fixation \textit{regions} generated by clustering the raw data and drawing an arbitrarily sized circle around the cluster center.}
	\label{fig:irises-cold}
\end{figure}

\subsection{Human-based attention maps}
We set up an experiment employing an eye tracker device to collect attention maps from human subjects who performed iris recognition task. Eye tracking enables following a person's gaze as he or she is looking around a screen, and calculating the numerical coordinates of the gaze with respect to the screen coordinate system, thus enabling a fairly precise analysis of what the user is looking at in any given moment. This is often used in psychological studies, marketing research and software usability studies, but the applications extend far beyond that, from OS navigation, gaming controls, to even enabling computer use for the severely disabled people. For the purpose of this study we have selected the EyeTribe device \cite{EyeTribe}. After a calibration procedure, the device outputs gaze coordinates in the form of $(x,y)$ points as a function of time, which can then be processed to come up with an attention map. These coordinates represent two types of gaze: fixations and saccades. Fixations occur when the person is currently looking at something, focusing the gaze on it. The opposite to fixations are saccades, constituting of larger eye movements, when the gaze in being moved between fixation points. As it has been proven that little to no visual processing cognition can be achieved during saccades \cite{EyeTrackingFixationsSaccades}, this allows focusing on the fixations periods when analyzing the gaze data, assuming that these periods contain the most useful information. Cluster analysis was then implemented on the raw data to find salient image regions by grouping together fixation points arranged similarly on the iris texture. These provided us regions that were used by human subjects during their comparison efforts, as depicted in Fig. \ref{fig:irises-cold}.

During this experiment, 28 subjects were asked to classify selected post-mortem iris image pairs as either genuine (same eye) or impostor (different eyes). Each subject could take as much time as they deemed necessary for coming up with their decision. The image pairs were randomly selected from the Warsaw-BioBase-PostMortem-Iris-v1 and v2 datasets, as shown in Fig. \ref{fig:irises-cold}. Since the GradCAM technique gives us the activation maps for the winning class, and our intent is to demonstrate and compare the \textit{correct} and \textit{incorrect} behaviors of the network, we evaluate the human-based attention maps from those pairs that were genuine as ground truth, but which were classified by humans as either genuine (\textit{correct}) or impostor (\textit{incorrect}).

\subsection{Decision accuracy vs post-mortem interval}
\begin{figure}[t]
	\centering
	\includegraphics[width=0.49\textwidth]{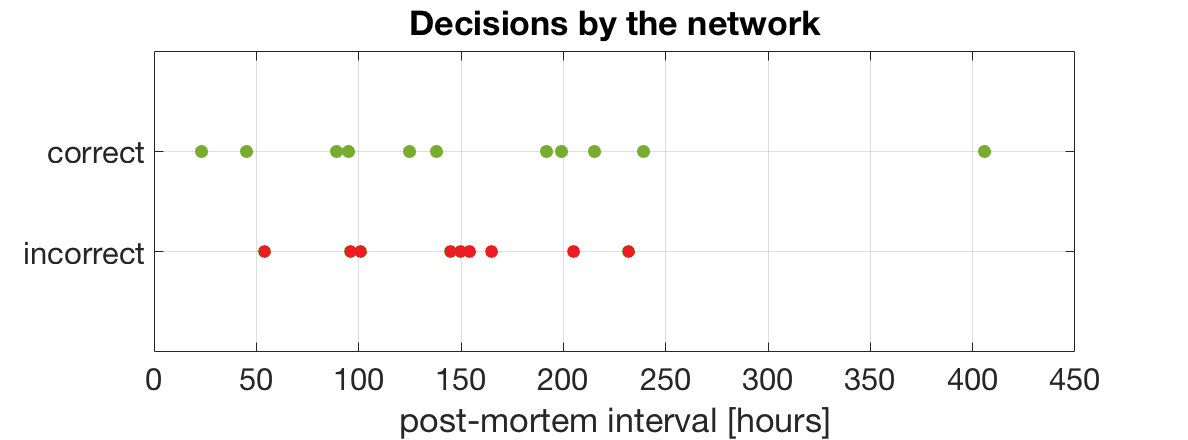}\\\vskip1mm
		\includegraphics[width=0.49\textwidth]{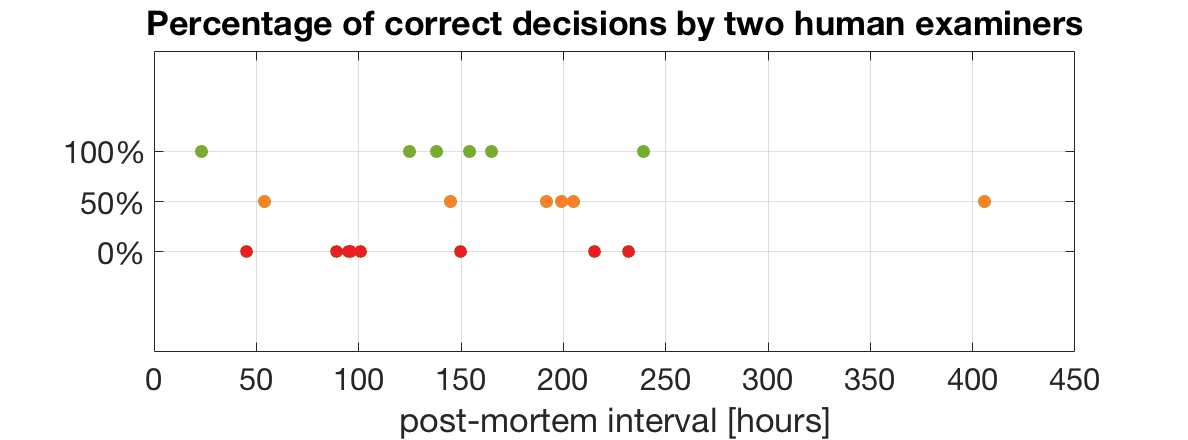}\\\vskip1mm
	\includegraphics[width=0.49\textwidth]{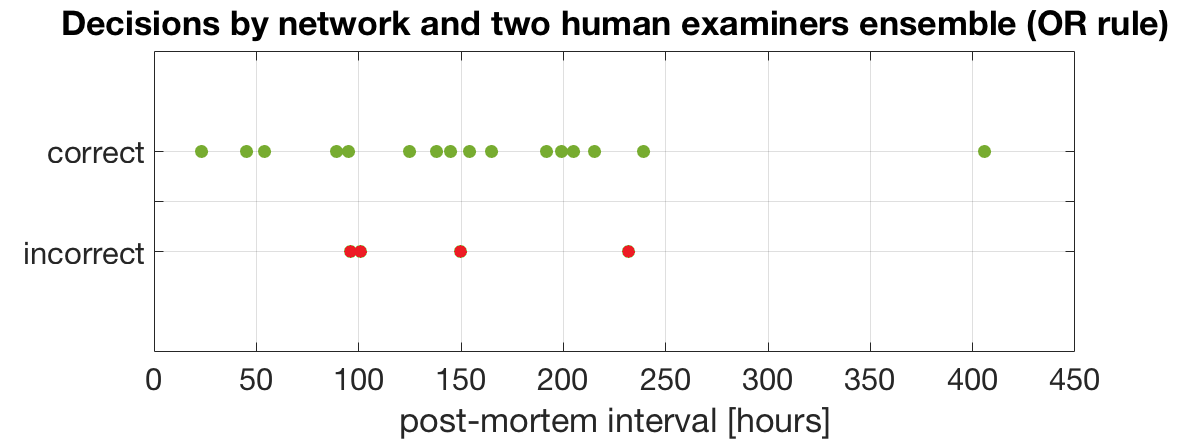}\\
	\caption{Decision accuracy achieved by DCNN, by pairs of humans, and by the ensemble of DCNN and the two humans, with respect to the post-mortem interval (PMI).}
	\label{fig:decisions_accuracy}
\end{figure}

Fig. \ref{fig:decisions_accuracy} presents the decision making accuracy achieved by the network, by pairs of human examiners (not necessarily the same), as well as by the ensemble of a machine solution and the two humans, with respect to the post-mortem interval (PMI). This is aggregated for the five cadaver eyes used in the attention map analysis described in the subsequent Section. Notably, there is no clear trend visible, \ie the longer PMI does not clearly contribute to lower decision making accuracy. Also, applying the OR rule to the machine and human decisions allowed to rectify most of the recognition errors, which may suggest that the machine classifier cold serve as an aid to the human expert.

\subsection{Human vs machine attention: a comparison}
\begin{figure}[!t]
	\begin{center}
	{\bf Similar maps:}\hskip22mm {\bf Different maps:}\\
		$q=0.318$ \hskip28mm $q=0.089$\\
	\includegraphics[width=0.112\textwidth]{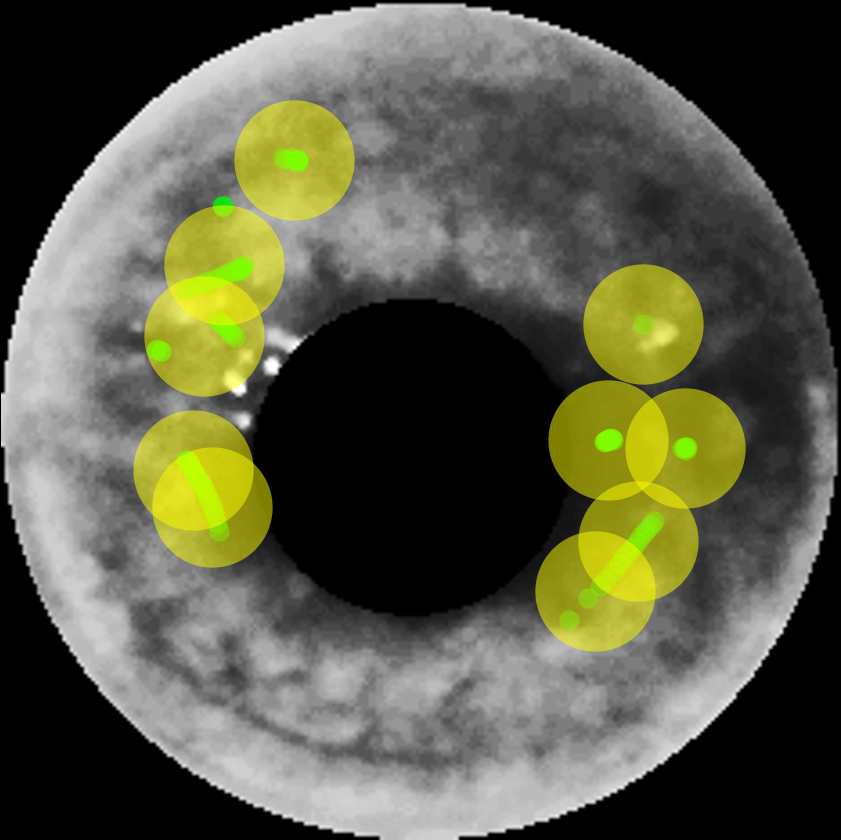}\hskip1mm
	\includegraphics[width=0.112\textwidth]{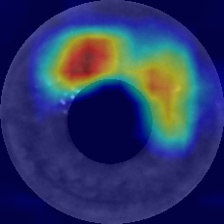}\hskip3mm
	\includegraphics[width=0.112\textwidth]{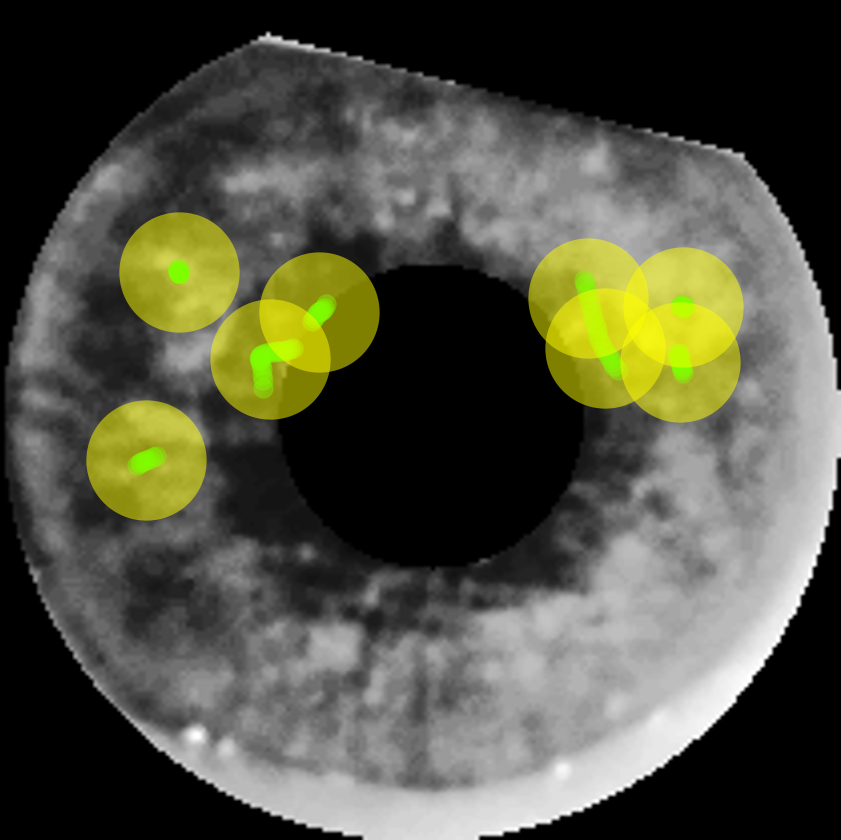}\hskip1mm
	\includegraphics[width=0.112\textwidth]{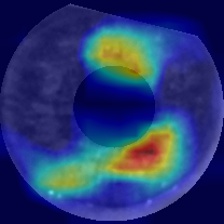}\\
		$q=0.191$ \hskip28mm $q=0.080$\\
	\includegraphics[width=0.112\textwidth]{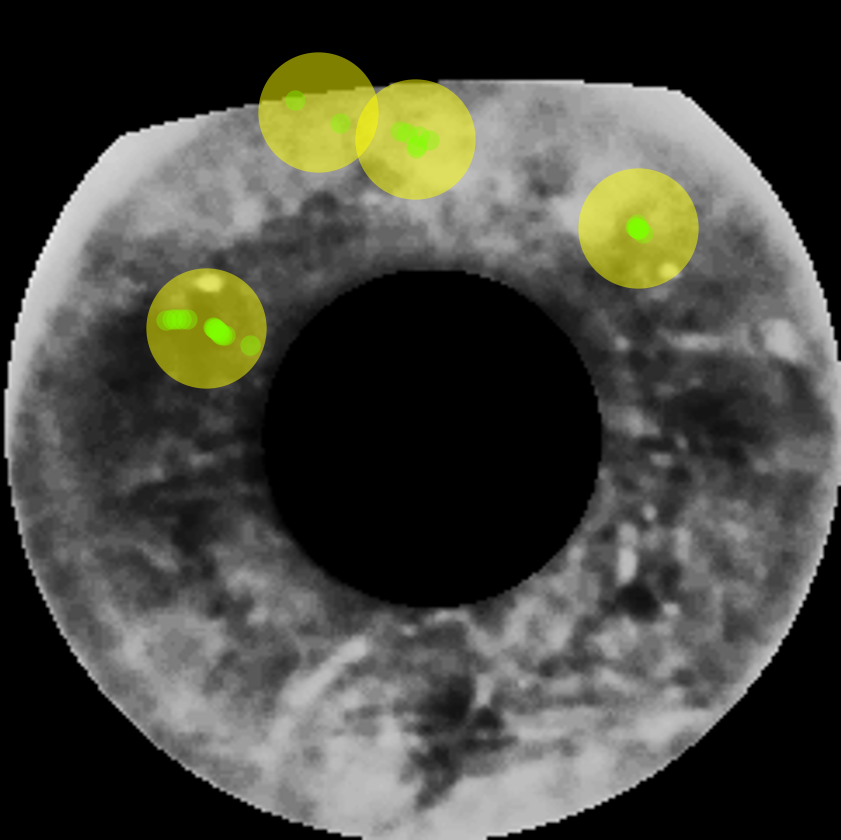}\hskip1mm
	\includegraphics[width=0.112\textwidth]{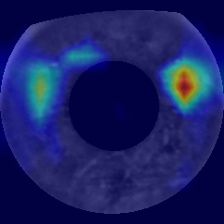}\hskip3mm
	\includegraphics[width=0.112\textwidth]{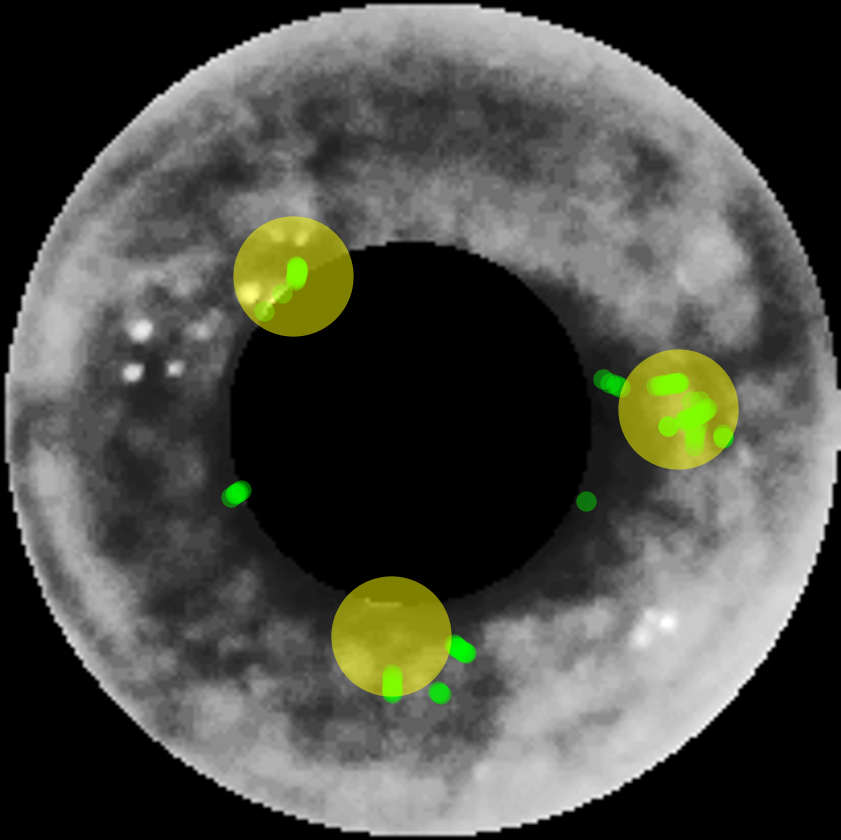}\hskip1mm
	\includegraphics[width=0.112\textwidth]{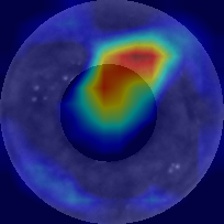}\\
		$q=0.221$ \hskip28mm $q=0.040$\\
	\includegraphics[width=0.112\textwidth]{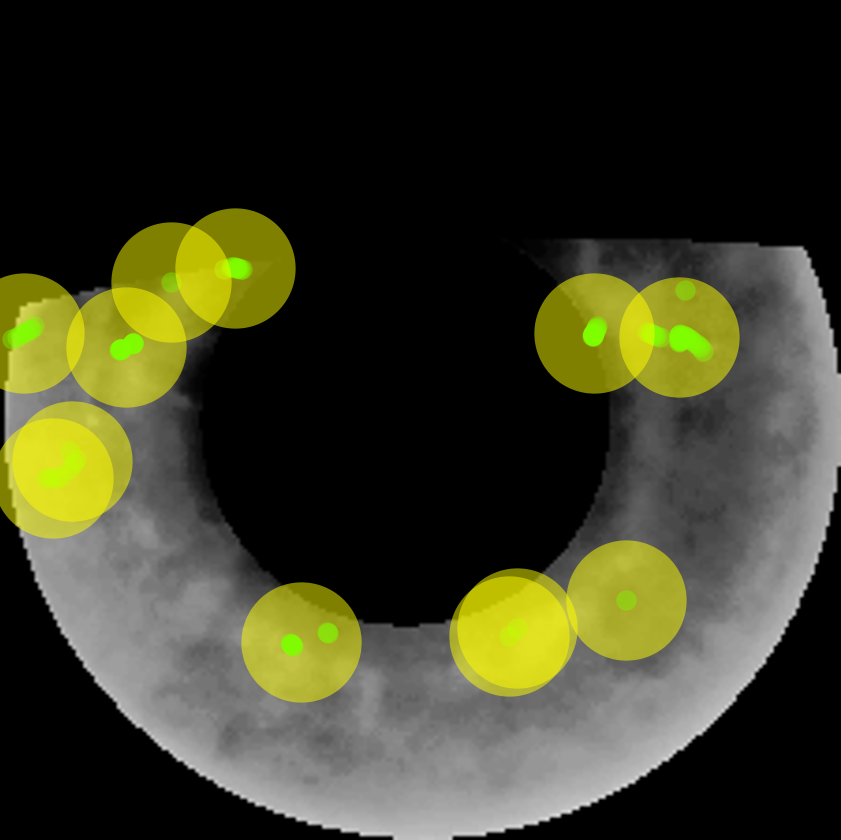}\hskip1mm
	\includegraphics[width=0.112\textwidth]{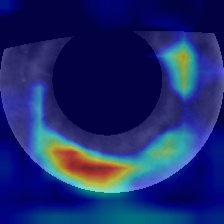}\hskip3mm
	\includegraphics[width=0.112\textwidth]{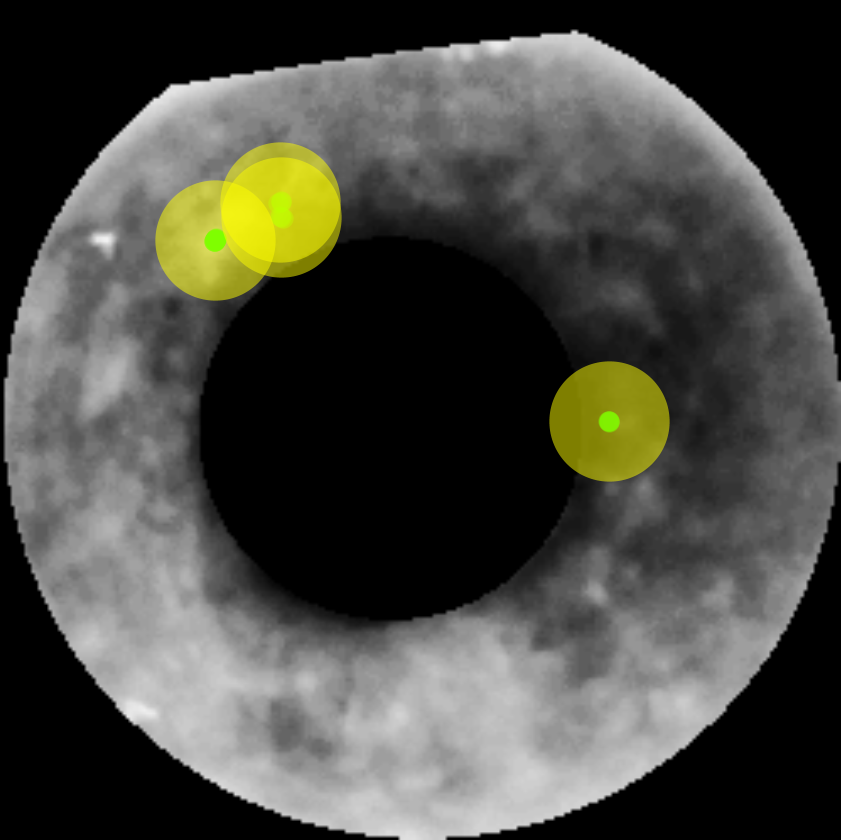}\hskip1mm
	\includegraphics[width=0.112\textwidth]{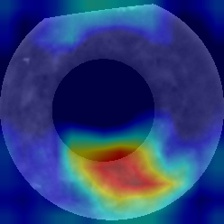}\\
	\end{center}
	\vskip-3mm
	\textcolor{green}{Correct}\hskip11mm
	\textcolor{red}{Incorrect}\hskip11mm
	\textcolor{green}{Correct}\hskip10mm
	\textcolor{red}{Incorrect}
	\caption{Human (gaze-tracking) and DCNN-based (CAM) attention maps when human subject provided a correct decision, but DCNN was wrong. Samples with similar maps are on the left, samples with dissimilar maps are on the right.}
	\label{fig:human_OK}
\end{figure}

\begin{figure}[!t]
	\begin{center}
		{\bf Similar maps:}\hskip22mm {\bf Different maps:}\\
		$q=0.216$ \hskip28mm $q=0.112$\\
	\includegraphics[width=0.112\textwidth]{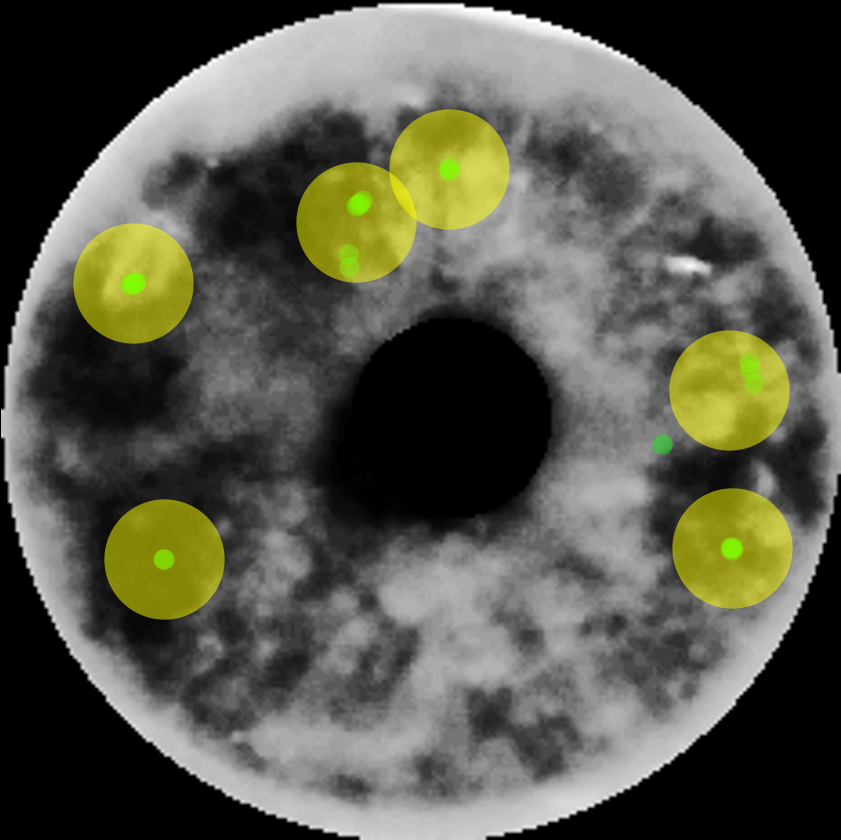}\hskip1mm
	\includegraphics[width=0.112\textwidth]{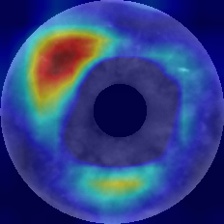}\hskip3mm
	\includegraphics[width=0.112\textwidth]{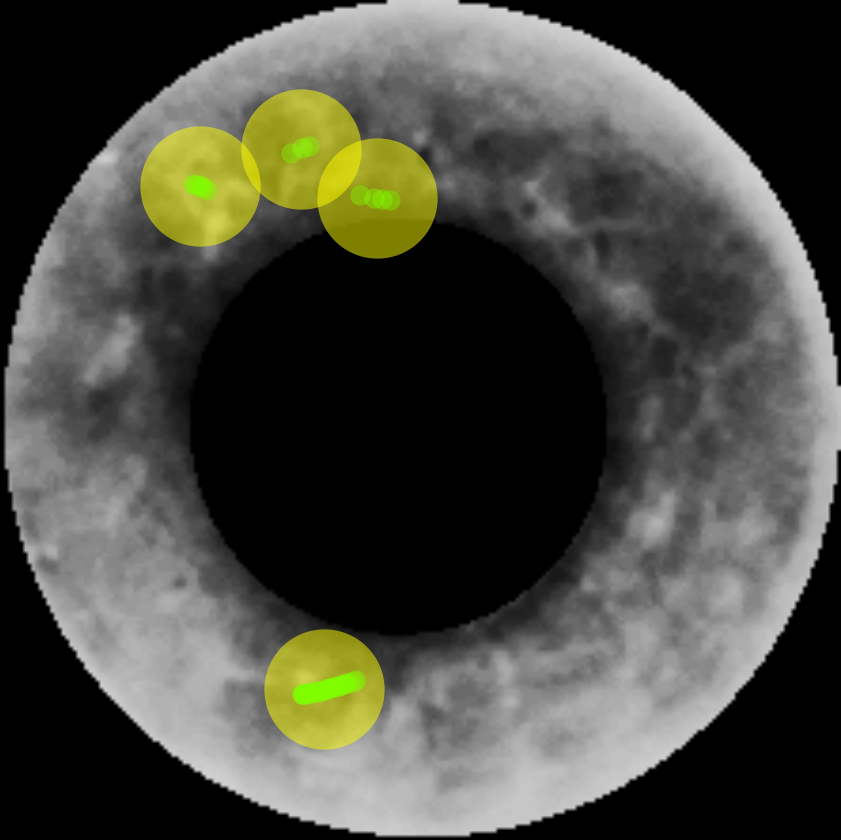}\hskip1mm
	\includegraphics[width=0.112\textwidth]{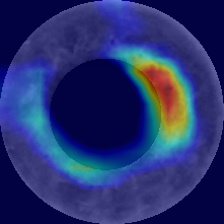}\\
		$q=0.259$ \hskip28mm $q=0.275$\\
	\includegraphics[width=0.112\textwidth]{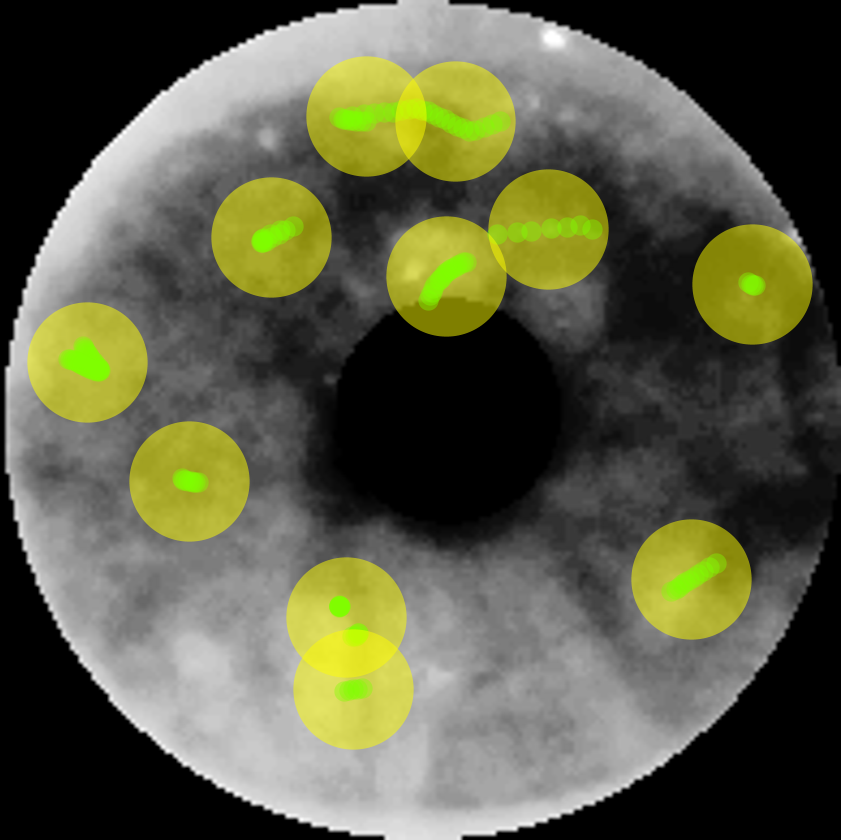}\hskip1mm
	\includegraphics[width=0.112\textwidth]{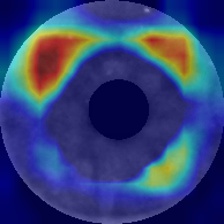}\hskip3mm
	\includegraphics[width=0.112\textwidth]{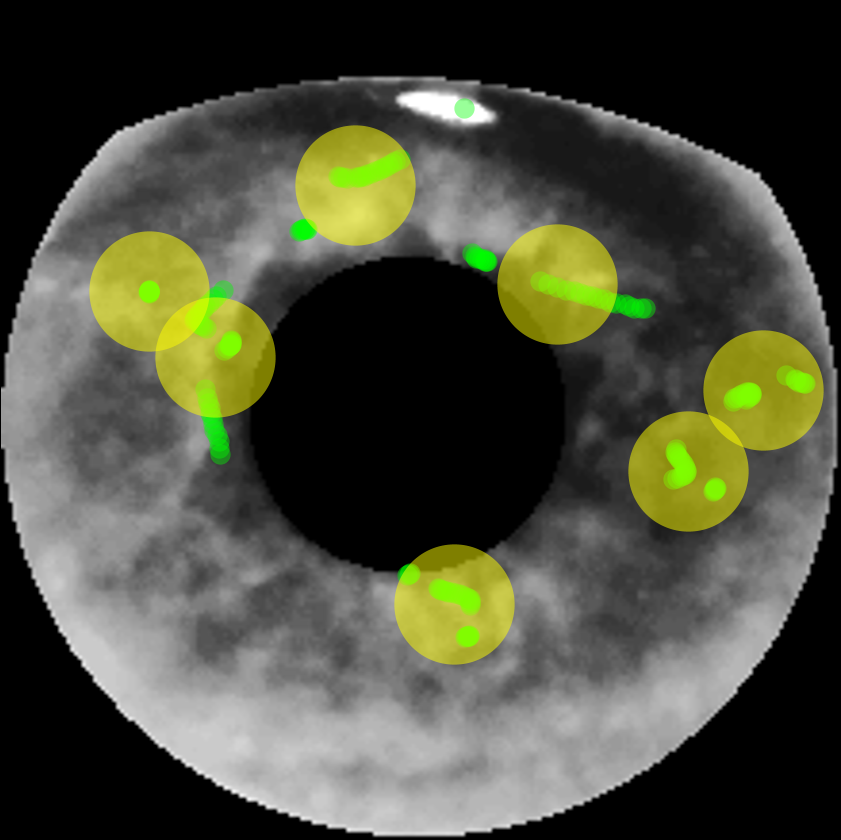}\hskip1mm
	\includegraphics[width=0.112\textwidth]{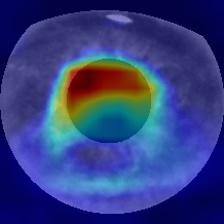}\\
		$q=0.068$ \hskip28mm $q=0.244$\\
	\includegraphics[width=0.112\textwidth]{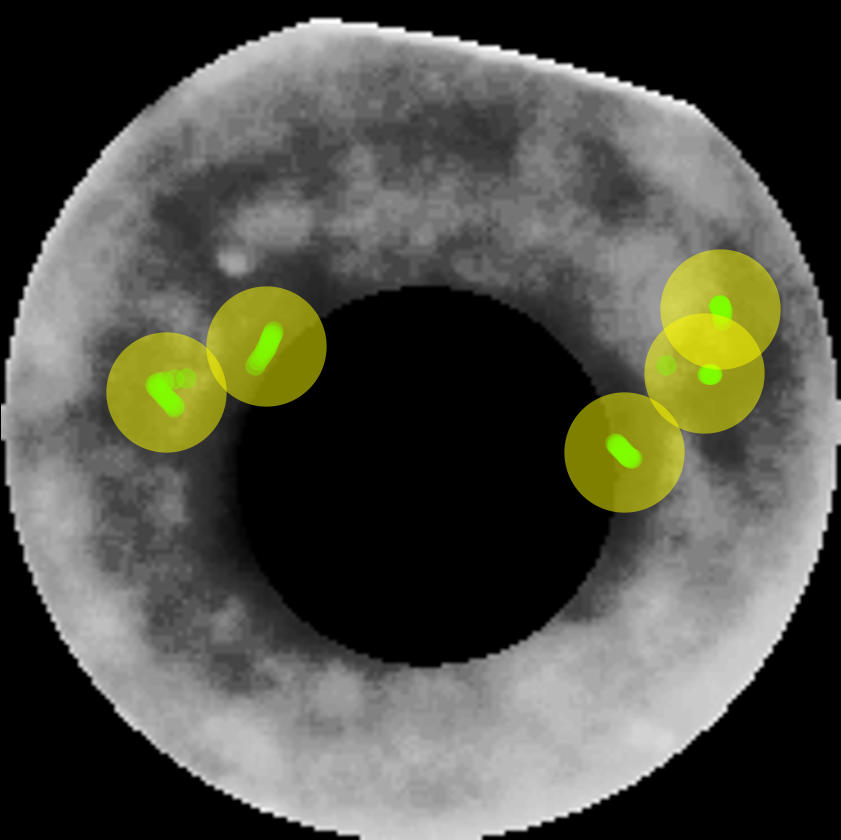}\hskip1mm
	\includegraphics[width=0.112\textwidth]{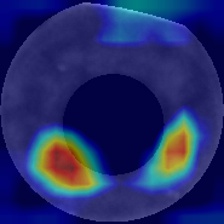}\hskip3mm
	\includegraphics[width=0.112\textwidth]{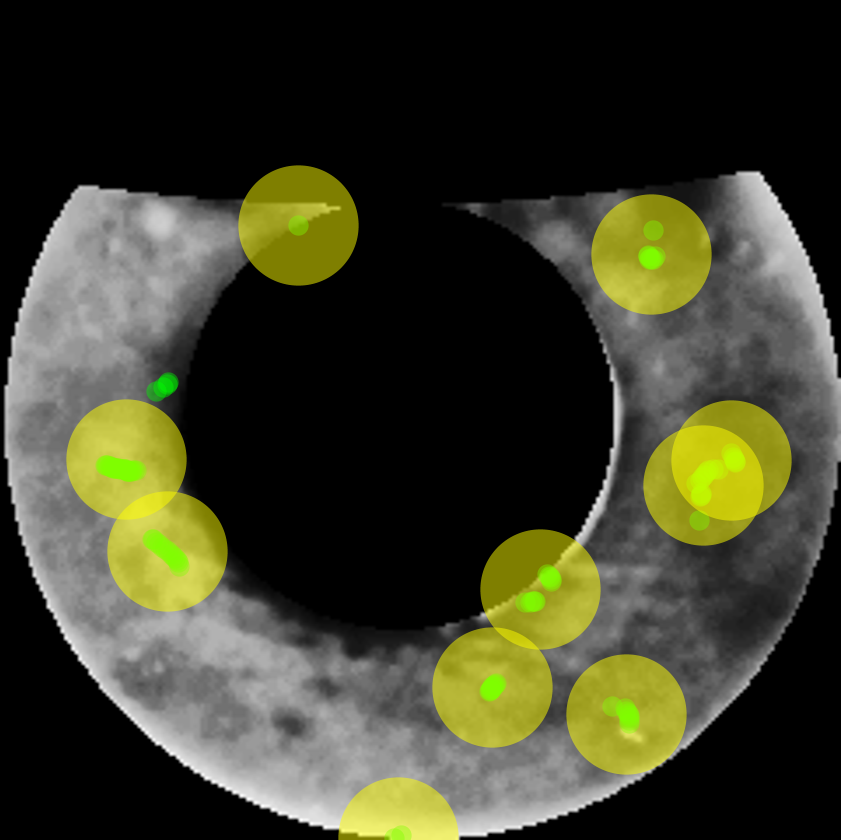}\hskip1mm
	\includegraphics[width=0.112\textwidth]{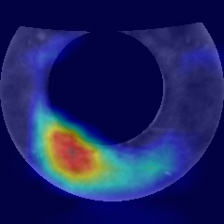}
	\end{center}
	\vskip-3mm
		\textcolor{green}{Correct}\hskip11mm
		\textcolor{green}{Correct}\hskip13mm
		\textcolor{green}{Correct}\hskip10mm
		\textcolor{green}{Correct}
	\caption{Cases when both the DCNN and human subject provided a correct decision.}
	\label{fig:both_OK}
\end{figure}

\begin{figure}[!t]
	\begin{center}
		{\bf Similar maps:}\hskip22mm {\bf Different maps:}\\
		$q=0.220$ \hskip28mm $q=0.181$\\
	\includegraphics[width=0.112\textwidth]{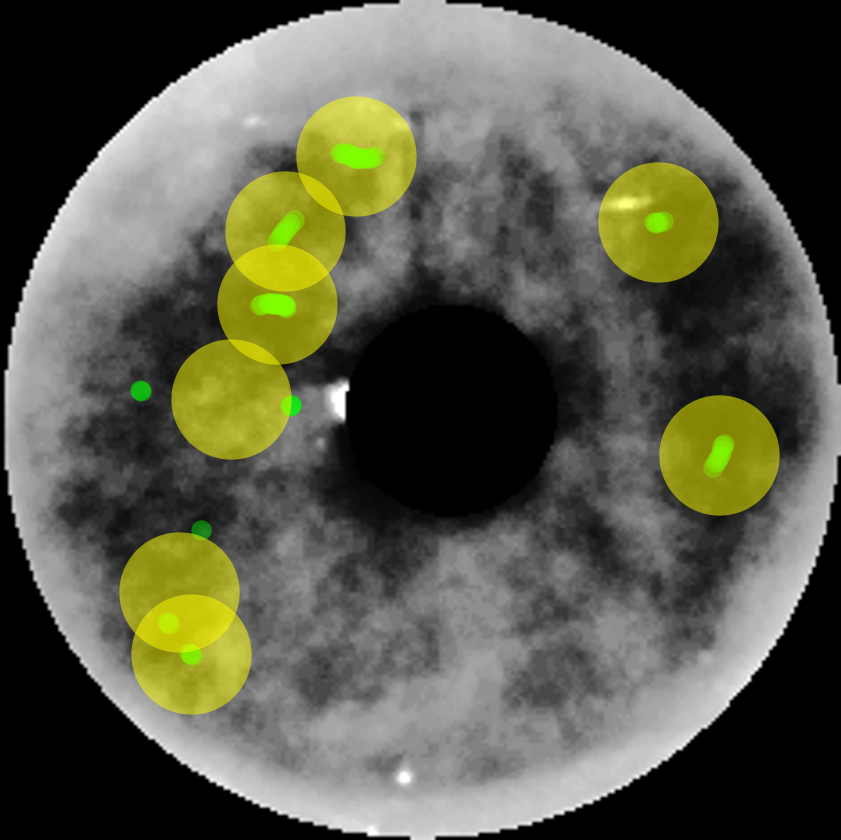}\hskip1mm
	\includegraphics[width=0.112\textwidth]{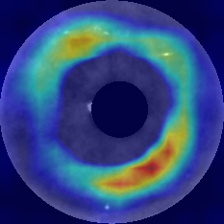}\hskip3mm
	\includegraphics[width=0.112\textwidth]{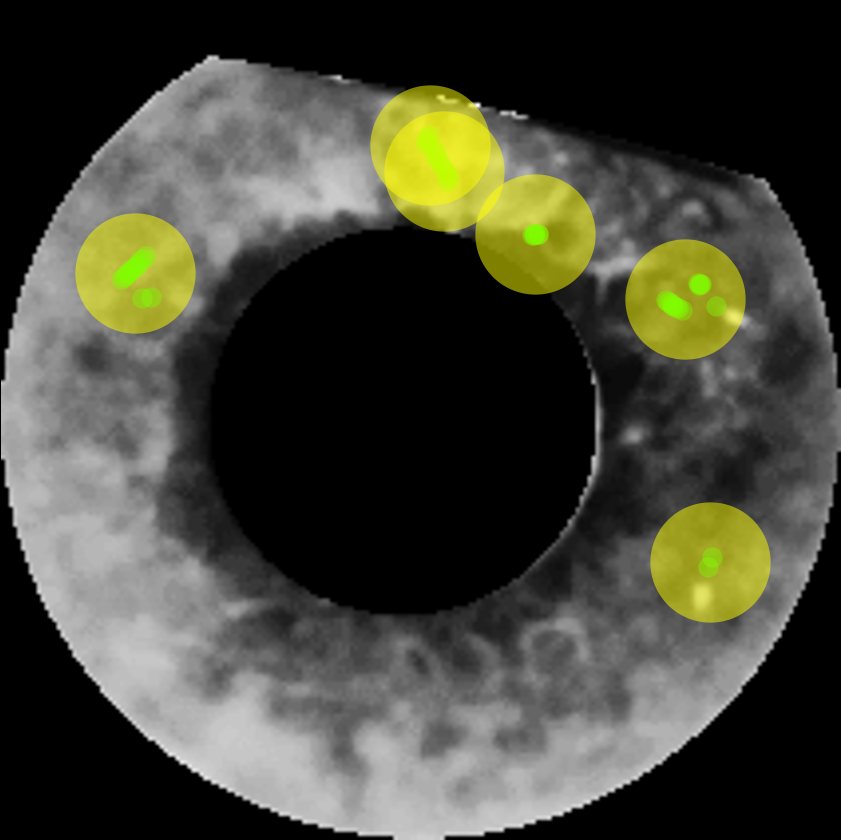}\hskip1mm
	\includegraphics[width=0.112\textwidth]{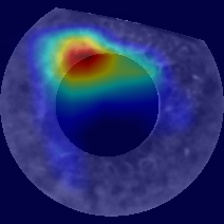}\\
		$q=0.153$ \hskip28mm $q=0.098$\\
	\includegraphics[width=0.112\textwidth]{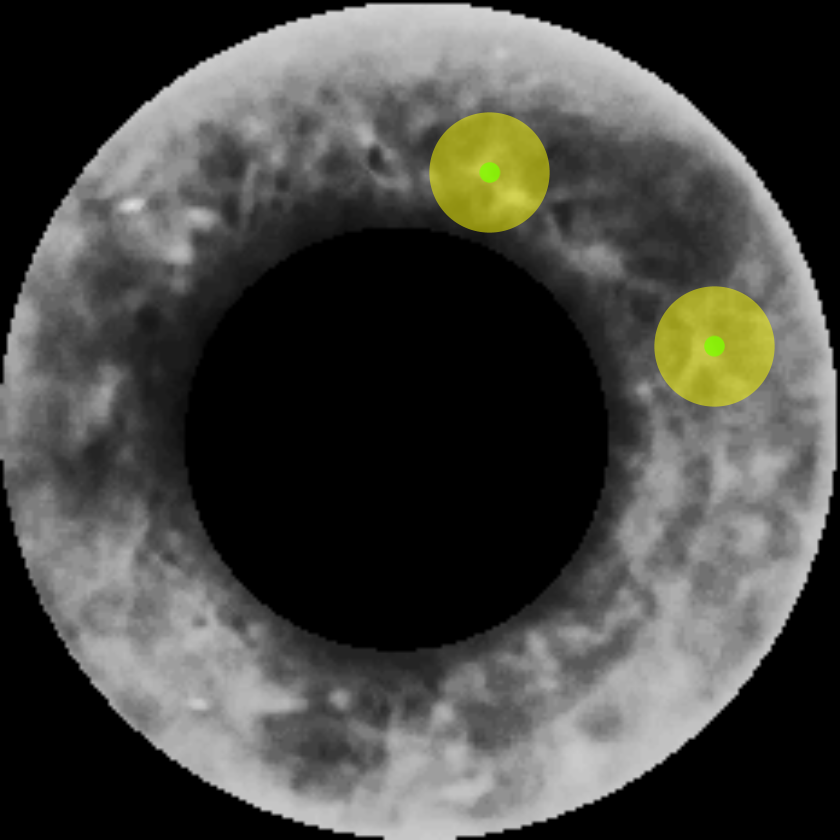}\hskip1mm
	\includegraphics[width=0.112\textwidth]{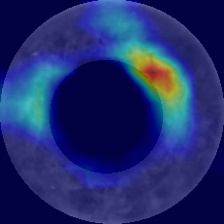}\hskip3mm
	\includegraphics[width=0.112\textwidth]{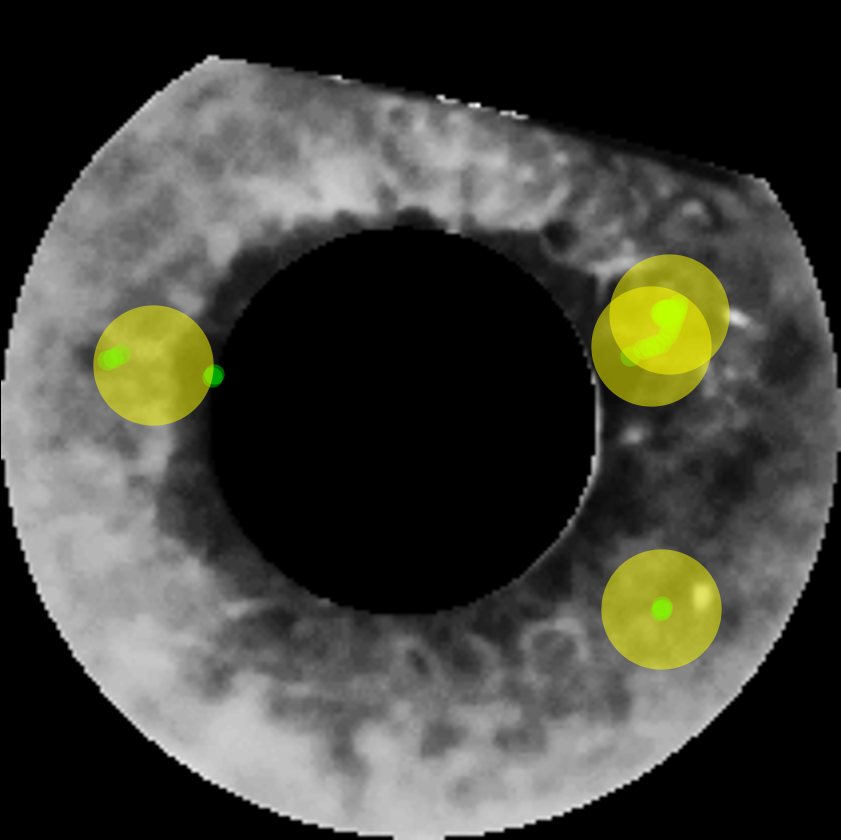}\hskip1mm
	\includegraphics[width=0.112\textwidth]{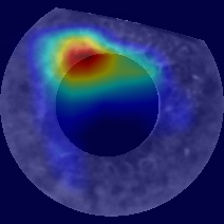}\\
		$q=0.189$ \hskip28mm $q=0.067$\\
	\includegraphics[width=0.112\textwidth]{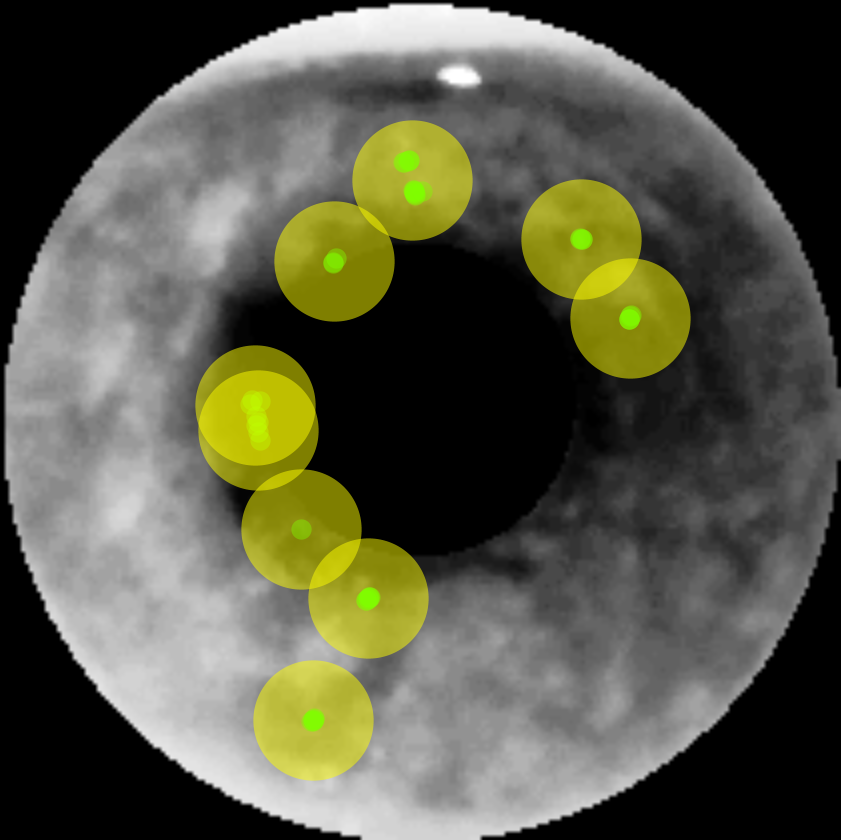}\hskip1mm
	\includegraphics[width=0.112\textwidth]{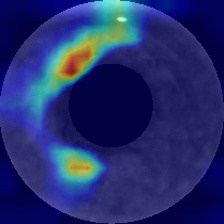}\hskip3mm
	\includegraphics[width=0.112\textwidth]{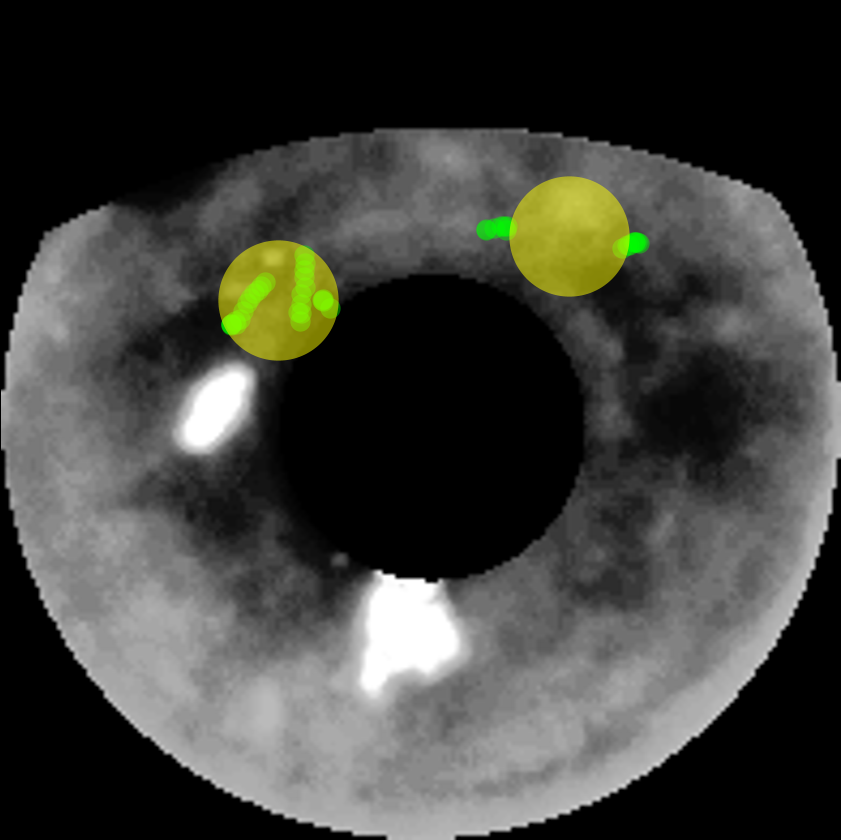}\hskip1mm
	\includegraphics[width=0.112\textwidth]{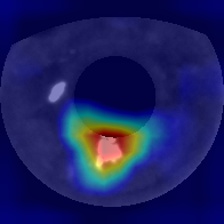}
	\end{center}
	\vskip-3mm
	\textcolor{red}{Incorrect}\hskip11mm
	\textcolor{green}{Correct}\hskip11mm
	\textcolor{red}{Incorrect}\hskip10mm
	\textcolor{green}{Correct}
	\caption{Cases when DCNN was correct, whereas the human subject was wrong.}
	\label{fig:DCNN_OK}
\end{figure}

\begin{figure}[!t]
	\begin{center}
		{\bf Similar maps:}\hskip22mm {\bf Different maps:}\\
	$q=0.198$ \hskip28mm $q=0.133$\\
	\includegraphics[width=0.112\textwidth]{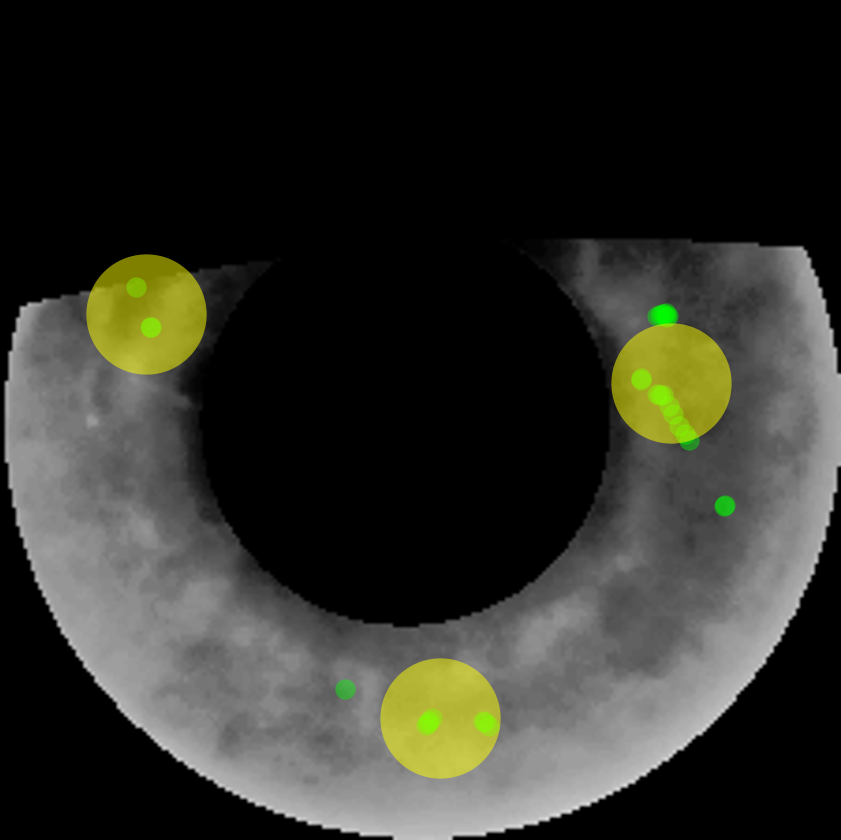}\hskip1mm
	\includegraphics[width=0.112\textwidth]{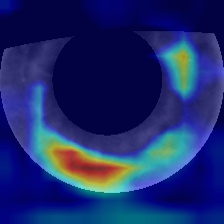}\hskip3mm
	\includegraphics[width=0.112\textwidth]{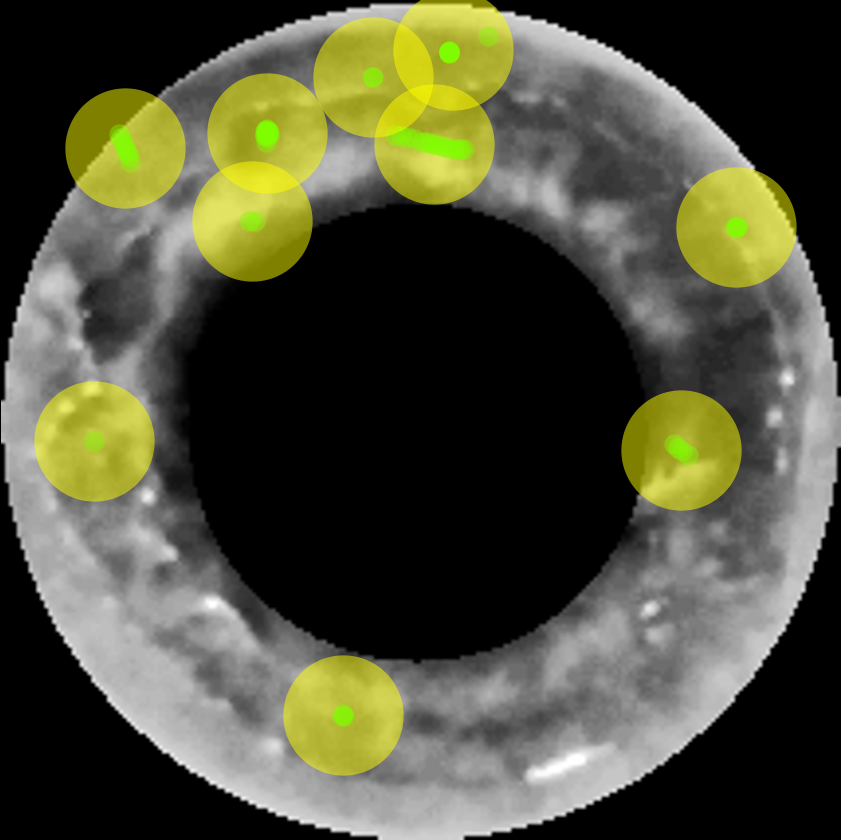}\hskip1mm
	\includegraphics[width=0.112\textwidth]{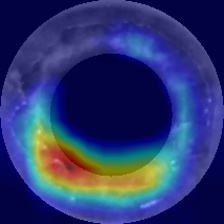}\\		$q=0.214$ \hskip28mm $q=0.058$\\
	\includegraphics[width=0.112\textwidth]{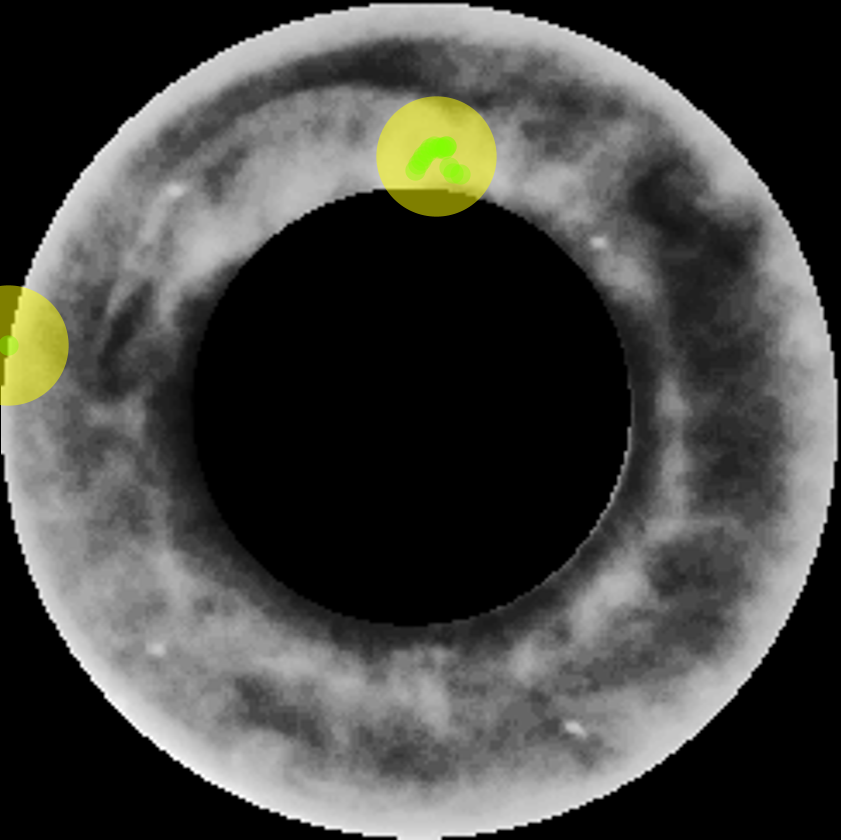}\hskip1mm
	\includegraphics[width=0.112\textwidth]{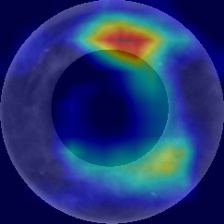}\hskip3mm
	\includegraphics[width=0.112\textwidth]{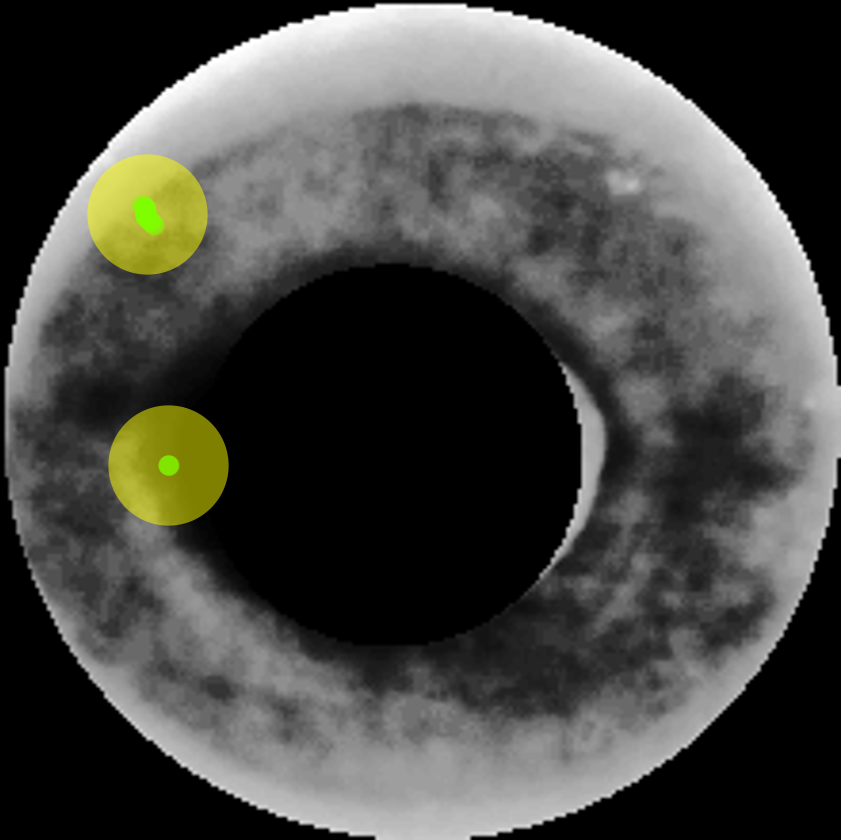}\hskip1mm
	\includegraphics[width=0.112\textwidth]{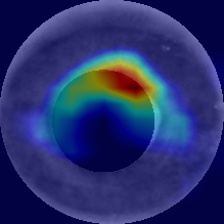}\\
		$q=0.071$ \hskip28mm $q=0.055$\\
	\includegraphics[width=0.112\textwidth]{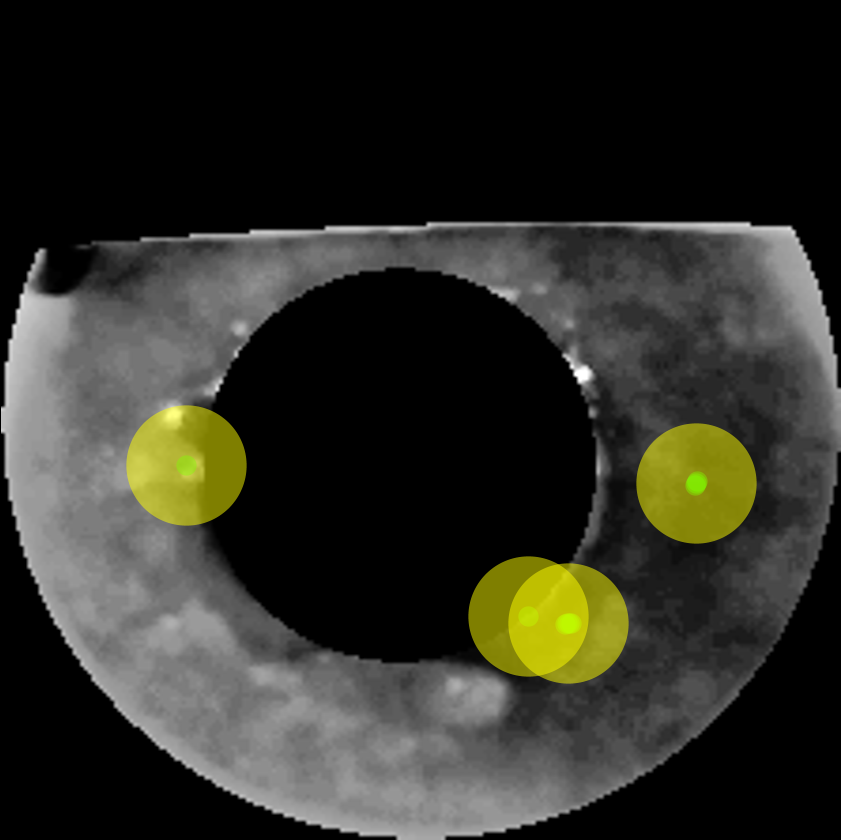}\hskip1mm
	\includegraphics[width=0.112\textwidth]{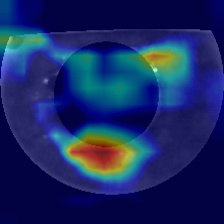}\hskip3mm
	\includegraphics[width=0.112\textwidth]{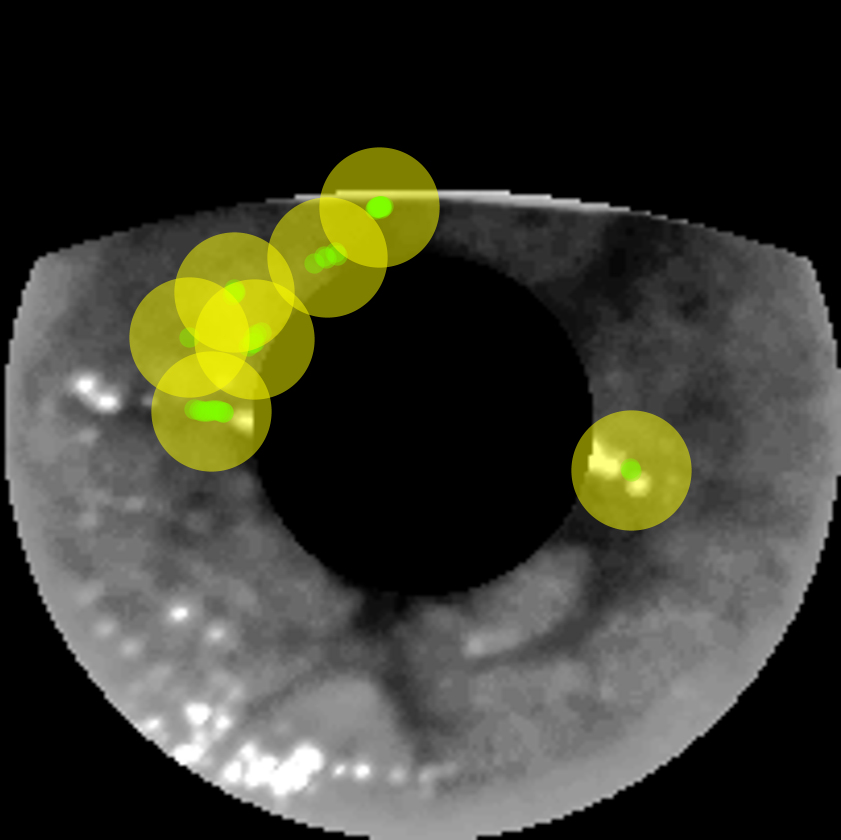}\hskip1mm
	\includegraphics[width=0.112\textwidth]{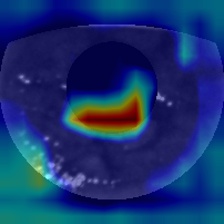}
	\end{center}
		\vskip-3mm
	\textcolor{red}{Incorrect}\hskip9mm
	\textcolor{red}{Incorrect}\hskip10.5mm
	\textcolor{red}{Incorrect}\hskip8mm
	\textcolor{red}{Incorrect}
	\caption{Cases when both DCNN and human subject provided an incorrect decision.}
	\label{fig:both_failed}
\end{figure}

In this Section, we present selected human-based attention maps, and compare them with class activation maps generated by the machine solution, similarly to those described in Sec. \ref{sec:CAM}, in four situations, namely:
\begin{itemize}
	\item when the DCNN misclassified a sample, but the human subject provided a correct decision, Fig. \ref{fig:DCNN_OK},
	\item when both the DCNN and the human subject provided a correct decision (same eye or different eyes on the presented pictures), Fig. \ref{fig:both_OK},
	\item when the human subject made a mistake, but the DCNN was correct, Fig. \ref{fig:human_OK},
	\item when both the human subject and the DCNN made a mistake, Fig. \ref{fig:both_failed}.
\end{itemize}

For each of the above, we consider two sub-cases: 1) when machine- and human-based attention maps are similar, and 2) when machine- and human-based attention maps point to different iris regions.

By inspecting these 8 cases in total, represented by 24 samples, we investigate the differences and similarities between human's and DCNN's attention to iris features, and see if the attention maps correspond to each other when the decision was correct, and when it was not. For the DCNN, a correct answer means giving the correct class-wise prediction. For experiments with human subjects, this means giving the correct genuine/impostor prediction.

Figure \ref{fig:human_OK} shows cases, in which the human subject gave the correct decision and the DCNN solution failed, despite attending the similar iris region as the human subject did (left pair). On the right, the DCNN also failed, but this time different attention maps are presented. Notably, both the machine and the human attended multiple iris regions, yet only human subject was able to give a correct answer.

Cases, for which the DCNN model gave correct answers are illustrated in Fig. \ref{fig:DCNN_OK}. On the top left, both the DCNN and the human subject are attending a large, circular region in the middle part of the iris. However, only the DCNN comes up with the correct solution. On the top right, the DCNN attends only a small portion of the iris, and still provides a correct answer, compared to the human subject, who fails despite attending more iris regions.

Fig. \ref{fig:both_OK} shows samples for which both the DCNN and the human subject were able to give correct decisions, supported by similar, and different attention maps.

Finally, in Fig. \ref{fig:both_failed} we show two samples for which both methods yielded incorrect results, supported by rather sparse (human subject on the left), but also by dense attention maps (DCNN on the left, human subject on the right).

\begin{figure*}[!t]
\centering
	\includegraphics[width=\textwidth]{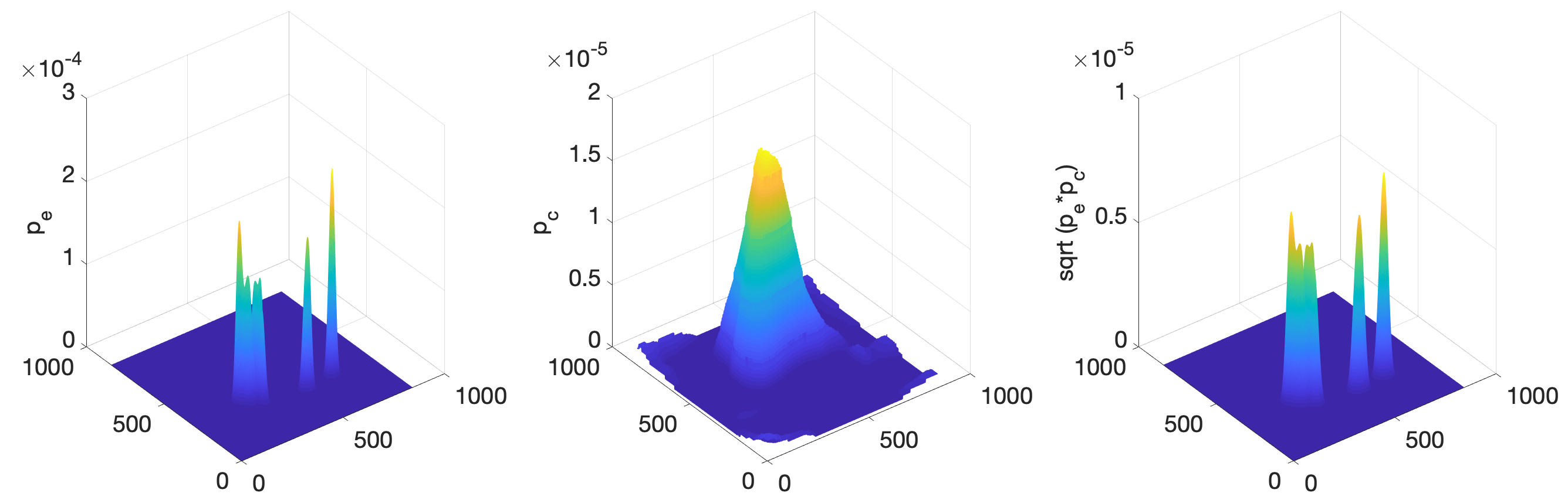}
	\caption{Estimation of salient regions obtained from eye-tracking ($p_e$, {\bf left}) and from CAM ($p_c$, {\bf middle}) for the same image. The square root of a product of $p_c$ and $p_e$  illustrates spatial agreement of salient regions between human and DCNN ({\bf right}).}
	\label{fig:probs}
\end{figure*}

In addition to qualitative (visual) assessment of correspondence of the DCNN and eye tracking-based salient regions, we provide a quantitative measure of how well these regions overlap as a geometric average $q$ of probability estimates $p_c$ for class activation maps and $p_e$ for eye tracking-based map:
\begin{equation}
	q = \sum_{i=1}^{N}\sum_{j=1}^{M}\sqrt{p_c(i,j)p_e(i,j)}
\end{equation}
where $\sum_{i,j}p_c(i,j) = 1$, $\sum_{i,j}p_e(i,j) = 1$, and $N,M$ determine the iris image size. Maps $p_c$ and $p_e$ may be interpreted as the probability of how important a given image region was for a network and for a human, respectively. While $p_c$ is a rescaled class activation map (to make a sum of elements equal to 1), the $p_e$ comes from fixation points convolved with an appropriate Gaussian function to account for the eye tracker uncertainty $\pm 20$ pixels for an HD screen ($1920\times1080$ pixels), as used in this work. Values of $q$ close to zero denote low overlap between human-driven and DCNN salient regions. Values of $q$ close to one denote very high level of agreement between the DCNN model and the human. These numerical assessments of the saliency regions similarity are given for each image pair in Figs. \ref{fig:human_OK} through \ref{fig:both_failed}. Fig. \ref{fig:probs} shows $p_c$, $p_e$ and the square root of their product for an example iris image.

Results described in this Section can be summarized in a few observations. First, although we were able to find samples for which both the DCNN-based and human-based attention maps were strikingly similar, we were as well able to find those that were region-disjoint. This suggests that DCNN-based visualization of salient iris regions may be complementary to what humans perceive as important in their judgements. Second, both the machine-based and human-based attention maps seem to omit the outer regions of the iris near the iris-sclera boundary, suggesting that discriminatory capacity of these areas for post-mortem iris samples may be limited.

\section{Discussion and Conclusions}
\label{sec:conclusions}
This study shows that despite the inherent difficulty found in the post-mortem iris image data, a DCNN-based classifier, fine-tuned to work with cadaver iris images, is able to efficiently learn discriminatory iris features and, when equipped with the class activation maps generation technique, can back its decisions in a human-intelligible way. The second deliverable of this study is the comparison between computer- and human-generated attention maps, with the latter being obtained with a gaze-tracking device.

These experiments are important in the sense that we are not aware of any other papers studying the human-based attention maps obtained during a gaze-tracking study with human asked to classify iris images as genuine or impostor, compared to what machines are doing. {\bf One conclusion} of this study is that appearance, similarity, or density of human-driven and machine-driven maps seem not to correspond in any clear way to decisions being made by either humans or machines. As for the similarities observed between humans and the neural network, both `examiners' tend to focus on a limited number of iris areas (often just one), which is opposite to the typically used iris code-based methods (such as Daugman's), analyzing the entire non-occluded portion of the iris annulus (sometimes additionally limited to ``non-fragile'' iris code bits). This may suggest that an effective way of post-mortem iris recognition may be based on sparse coding (such as minutiae-based coding in fingerprints, or keypoint-based object recognition) rather than on dense, iris code-based algorithms. The {\bf second conclusion} is that both humans and DCNN focused more on the inner/middle part of the iris, what suggests that outer parts (close to sclera) may be less effective in post-mortem iris recognition. {\bf The third conclusion} from this work is that salient regions proposed by the DCNN and identified from human eye gaze do not overlap in general, hence the computer-added visual cues may potentially constitute a valuable addition to the forensic examiner's expertise, as it can highlight important discriminatory regions that the human expert might miss in their proceedings. {\bf The fourth conclusion} from this study is that human subjects can provide an incorrect decision even despite spending quite some time observing many iris regions. Thus, we may hazard a guess that iris features `extracted' by non-expert human subjects do not always allow for post-mortem iris recognition, and an additional training may be necessary, similar to the training of forensic experts dealing with fingerprint, to become effective in recognizing post-mortem iris images.

\begin{thebibliography}{10}\itemsep=-1pt

\bibitem{BostonPostMortem}
{A. Sansola}.
\newblock {Postmortem iris recognition and its application in human
  identification}, MSc Thesis, Boston Univ., 2015.

\bibitem{BolmeBTAS2016}
D.~S. Bolme et~al.
\newblock Impact of environmental factors on biometric matching during human
  decomposition.
\newblock {\em IEEE 8th International Conference on Biometrics Theory,
  Applications and Systems, Sep 6-9, 2016, Buffalo, USA}, 2016.

\bibitem{Keras}
F.~Chollet.
\newblock {Keras: Deep Learning library for Theano and TensorFlow}.
\newblock 2015.

\bibitem{Daugman1993}
J.~Daugman.
\newblock High confidence visual recognition of persons by a test of
  statistical independence.
\newblock {\em IEEE Transactions on Pattern Analysis and Machine Intelligence},
  15(11):1148--1161, Nov 1993.

\bibitem{Daugman2007NewMethods}
J.~Daugman.
\newblock New methods in iris recognition.
\newblock {\em IEEE Transactions on Systems, Man, and Cybernetics -- Part B:
  Cybernetics}, 37(5):1167--1175, 2007.

\bibitem{DeepIrisNet2016Gangwar}
A.~Gangwar and A.~Joshi.
\newblock {DeepIrisNet: Deep iris representation with applications in iris
  recognition and cross-sensor iris recognition}.
\newblock {\em IEEE International Conference on Image Processing (ICIP 2016)},
  2016.

\bibitem{Hollingsworth_BTAS_2009}
K.~P. Hollingsworth, K.~W. Bowyer, and P.~J. Flynn.
\newblock Using fragile bit coincidence to improve iris recognition.
\newblock {\em Proceedings of the 3rd IEEE International Conference on
  Biometrics: Theory, Applications and Systems}, 2009.

\bibitem{DeepIrisLIU2016}
N.~Liu, M.~Zhang, H.~Li, Z.~Sun, and T.~Tan.
\newblock {DeepIris: Learning pairwise filter bank for heterogeneous iris
  verification}.
\newblock {\em Pattern Recognition Letters}, 82:154 -- 161, 2016.

\bibitem{tensorflow2015}
{M. Abadi \emph{et al.}}
\newblock {TensorFlow: Large-Scale Machine Learning on Heterogeneous Systems},
  2015.
\newblock tensorflow.org.

\bibitem{MinaeeCNNforIrisRec2016}
S.~Minaee, A.~Abdolrashidiy, and Y.~Wang.
\newblock {An Experimental Study of Deep Convolutional Features For Iris
  Recognition}.
\newblock {\em IEEE Signal Processing in Medicine and Biology Symposium (SPMB
  2016)}, 2016.

\bibitem{IrisCNNsOffTheShelfNguyen2018}
K.~Nguyen, C.~Fookes, A.~Ross, and S.~Sridharan.
\newblock {Iris Recognition With Off-the-Shelf CNN Features: A Deep Learning
  Perspective}.
\newblock {\em IEEE Access}, 6:848 -- 855, 2018.

\bibitem{GradCAMCode}
V.~Petsiuk.
\newblock {Keras implementation of GradCAM}.
\newblock accessed: April 4, 2018.

\bibitem{RossPostMortem}
A.~Ross.
\newblock {Iris as a Forensic Modality: The Path Forward}.

\bibitem{EyeTrackingFixationsSaccades}
D.~Salvucci and J.~Goldberg.
\newblock Identifying fixations and saccades in eye-tracking protocols.
\newblock In {\em Eye Tracking Research and Applications Symposium}, pages
  71--78, 2000.

\bibitem{PostMortemPigs}
S.~K. Saripalle, A.~McLaughlin, R.~Krishna, A.~Ross, and R.~Derakhshani.
\newblock {Post-mortem Iris Biometric Analysis in Sus scrofa domesticus}.
\newblock {\em IEEE 7th International Conference on Biometrics Theory,
  Applications and Systems (BTAS 2015), September 8-11, 2015, Arlington, USA},
  2015.

\bibitem{Sauerwein_JFO_2017}
K.~Sauerwein, T.~B. Saul, D.~W. Steadman, and C.~B. Boehnen.
\newblock The effect of decomposition on the efficacy of biometrics for
  positive identification.
\newblock {\em Journal of Forensic Sciences}, 62(6):1599--1602, 2017.

\bibitem{GradCAMSelvaraju}
R.~R. Selvaraju, M.~Cogswell, A.~Das, R.~Vedantam, D.~Parikh, and D.~Batra.
\newblock {Grad-CAM: Visual Explanations from Deep Networks via Gradient-Based
  Localization}.
\newblock {\em International Conference on Computer Vision (ICCV)}, 2016.

\bibitem{VGGSimonyanCNNsForRecognition2014}
K.~Simonyan and A.~Zisserman.
\newblock {Very Deep Convolutional Networks for Large-Scale Image Recognition}.
\newblock {\em International Conference on Learning Representations}, 2014.

\bibitem{EyeTribe}
{TheEyeTribe}.
\newblock {The EyeTribe Documentation and API Reference},
  https://github.com/EyeTribe/documentation\#category-tracker (accessed:
  November 20, 2017).

\bibitem{TrokielewiczIWBF2018}
M.~Trokielewicz and A.~Czajka.
\newblock {Data-driven Segmentation of Post-mortem Iris Images}.
\newblock {\em Intl Workshop on Biometrics and Forensics (IWBF2018), June 7-8,
  2018, Sassari, Italy}.

\bibitem{TrokielewiczPostMortemBTAS2016}
M.~Trokielewicz, A.~Czajka, and P.~Maciejewicz.
\newblock {Human Iris Recognition in Post-mortem Subjects: Study and Database}.
\newblock {\em 8th IEEE International Conference on Biometrics: Theory,
  Applications and Systems, Sep 6-9, 2016, Buffalo, USA}.

\bibitem{TrokielewiczPostMortemICB2016}
M.~Trokielewicz, A.~Czajka, and P.~Maciejewicz.
\newblock {Post-mortem Human Iris Recognition}.
\newblock {\em 9th IAPR International Conference on Biometrics (ICB 2016), June
  13-16, 2016, Halmstad, Sweden}, 2016.

\bibitem{TrokielewiczTIFS2018}
M.~Trokielewicz, A.~Czajka, and P.~Maciejewicz.
\newblock Iris recognition after death.
\newblock {\em IEEE Transactions on Information Forensics and Security},
  14(6):1501--1514, 2018.

\bibitem{TrokielewiczColdPAD_BTAS2018}
M.~Trokielewicz, A.~Czajka, and P.~Maciejewicz.
\newblock {Presentation Attack Detection for Cadaver Iris}.
\newblock {\em 9th IEEE International Conference on Biometrics: Theory,
  Applications and Systems, Oct 22-25, 2018, Los Angeles, USA}, 2018.

\bibitem{ZhaoDeepIrisICCV2017}
Z.~Zhao and A.~Kumar.
\newblock {Towards More Accurate Iris Recognition Using Deeply Learned
  Spatially Corresponding Features}.
\newblock {\em IEEE International Conference on Computer Vision (ICCV 2017)},
  2017.

\bibitem{CAMZhou}
B.~Zhou, A.~Khosla, A.~Lapedriza, A.~Oliva, and A.~Torralba.
\newblock {Learning Deep Features for Discriminative Localization}.
\newblock {\em International IEEE Conference on Computer Vision and Pattern
  Recognition (CVPR 2016)}, 2016.

\end{thebibliography}
\end{document}